\documentclass[twoside,11pt]{article}

\usepackage{blindtext}
\usepackage{xcolor}

\usepackage{jmlr2e}
\usepackage{amsmath,amssymb,amsfonts}
\usepackage{algorithm}
\usepackage{algpseudocode}
\usepackage{ulem}
\usepackage{graphicx}
\usepackage{textcomp}
\def\BibTeX{{\rm B\kern-.05em{\sc i\kern-.025em b}\kern-.08em
    T\kern-.1667em\lower.7ex\hbox{E}\kern-.125emX}}
\DeclareMathOperator*{\argmin}{arg\,min}
\DeclareMathOperator*{\argmax}{arg\,max}
\usepackage{pgfgantt}
\usepackage{subcaption}
\usepackage{multirow}
\tikzstyle{arrow} = [thick,->,>=stealth]
\tikzset{
  labelnodes/.style={text width=6cm,align=right,font=\RaggedLeft, left=7pt,fill=blue!20,}
}
\ganttset{bar incomplete/.append style={fill=red!40},
    group/.append style={draw=black, fill=red},}
    
\usepackage{lastpage}

\ShortHeadings{ICDARTS: Improving the Stability and Expanding the Search Space of Cyclic DARTS}{Herron, Rose, and Young}
\firstpageno{1}

\begin{document}

\title{ICDARTS: Improving the Stability and Performance of Cyclic DARTS}

\author{\name Emily Herron \email eherron5@vols.utk.edu \\
       \addr Bredesen Center for Interdisciplinary \\
       Research and Graduate Education\\
       University of Tennessee\\
       Knoville, TN 37996-3394, USA
       \AND
       \name Derek Rose \email rosedc@ornl.gov \\
       \addr Electrification and Energy Infrastructure Division\\
       Oak Ridge National Laboratory\\
       Oak Ridge, TN 37831-6075, USA
       \AND
       \name Steven Young \email youngsr@ornl.gov \\
       \addr Computer Science and Mathematics Division\\
       Oak Ridge National Laboratory\\
       Oak Ridge, TN 37831-6164, USA}

\editor{}

\maketitle

\begin{abstract}
NOTE: This is an expanded version of a previously published conference paper \citep{herron2022icdarts}. This paper includes an expanded study of the importance of each algorithm change, an ablation study of the importance of each layer choice, a study of the effect of different layer choices, and a study of performing ICDARTS NAS on a dynamic search space.

This work introduces improvements to the stability and generalizability of Cyclic DARTS (CDARTS). CDARTS is a Differentiable Architecture Search (DARTS)-based approach to neural architecture search (NAS) that uses a cyclic feedback mechanism to train search and evaluation networks concurrently. This training protocol aims to optimize the search process by enforcing that the search and evaluation networks produce similar outputs. However, CDARTS introduces a loss function for the evaluation network that is dependent on the search network. The dissimilarity between the loss functions used by the evaluation networks during the search and retraining phases results in a search-phase evaluation network that is a sub-optimal proxy for the final evaluation network that is utilized during retraining. We present ICDARTS, a revised approach that eliminates the dependency of the evaluation network weights upon those of the search network, along with a modified process for discretizing the search network's \textit{zero} operations that allows these operations to be retained in the final evaluation networks. We pair the results of these changes with ablation studies on ICDARTS' algorithm and network template. Finally, we explore methods for expanding the search space of ICDARTS by expanding its operation set and exploring alternate methods for discretizing its continuous search cells. These experiments resulted in networks with improved generalizability and the implementation of a novel method for incorporating a dynamic search space into ICDARTS.  
\end{abstract}

\footnote{Notice: This manuscript has been authored by UT-Battelle, LLC, under contract DE-AC05-00OR22725 with the US Department of Energy (DOE). The US government retains and the publisher, by accepting the article for publication, acknowledges that the US government retains a nonexclusive, paid-up, irrevocable, worldwide license to publish or reproduce the published form of this manuscript, or allow others to do so, for US government purposes. DOE will provide public access to these results of federally sponsored research in accordance with the DOE Public Access Plan (https://www.energy.gov/doe-public-access-plan).}

\begin{keywords}
 Neural Architecture Search, AutoML, Deep Learning, Neural Networks
\end{keywords}

\section{Introduction}
In deep learning, deep neural networks are applied to a range of vision tasks, including image recognition \citep{tan2021efficientnetv2}, object detection \citep{Redmon_2017_CVPR,He_2017_ICCV,Wang_2020_CVPR}, and semantic segmentation \citep{yu2018methods,nekrasov2019fast}. When encountering a new task, machine learning researchers often deploy a state-of-the-art deep learning architecture designed for and evaluated on a limited number of popular benchmark datasets. Since image datasets can vary widely in size, resolution, and subject matter, a single state-of-the-art network architecture that generalizes well to one task may perform poorly on a different dataset.
NAS algorithms strive to design optimal network architectures for new tasks and datasets efficiently.
Some popular neural architecture search algorithms rely on reinforcement learning \citep{zoph2017neural,zoph2018learning}, evolutionary algorithms \citep{xie2017genetic,real2019regularized,elsken2019efficient,yang2020cars}, and gradient optimization \citep{kandasamy2019neural,liu2018hierarchical}. 
One gradient-based algorithm that has become prominent in NAS is DARTS \citep{liu2019darts}.
In a departure from previous methods that rely on discrete search spaces, DARTS represents its search space as a set of continuous and differentiable directed acyclic graph structures called cells. 
Architectures are learned in the algorithm's initial search phase by optimizing the continuous connections between cell nodes and their associated operations using a gradient-based approach. 
Then, in a separate evaluation phase, the cells are discretized by removing all but the most optimal connections at each depth. A final, deeper architecture is finally constructed from these discretized cells and evaluated on the full dataset.

The popularity of the original DARTS paper inspired multiple follow-up publications. One presents the Progressive DARTS algorithm (P-DARTS) \citep{chen2019progressive}, which addresses the discrepancy between the networks' depths during DARTS' search and evaluation phases by gradually increasing the depth of the network during the search phase. Another publication introduces the CDARTS \citep{yu2020cyclic} method, which builds upon P-DARTS by proposing a cyclic feedback mechanism between the search and evaluation networks that allows the networks to be trained jointly in a cyclic manner. This joint learning strategy ensures that the search and evaluation networks are each optimized while learning similar features and producing similar results.

Although CDARTS begins to address the issue of DARTS' continuous search network by imperfectly predicting the performance of the discretized network, we find that its current method for training its evaluation network in the search phase is inconsistent with the method that was used for retraining this network in the evaluation phase. This discrepancy decreases the value of this network as a proxy for the final discretized network.
We contribute improvements to the algorithm and search space of CDARTS that result in improved stability and generalization ability. Our enhanced version of the CDARTS algorithm, ICDARTS, presents a training procedure for the evaluation network at search time that better resembles its retraining process. We pair the results of these improvements with two ablation studies: the first on the elements of the CDARTS template, including the layer type options, reduce cells, and stemming layers, and the second on two routes of incremental changes that transform the CDARTS algorithm to ICDARTS. In addition to our algorithmic changes, we found that modifying and expanding the default search space of ICDARTS further improved the algorithm's stability and generalization ability. The first of these changes involved eliminating the \textit{zero} operation from the operation of ICDARTS since this operation is removed upon discretization in the original version of CDARTS and is thus irrelevant to the final network. Next, we explored alternative sets of operation choices of varying complexity for ICDARTS. These experiments culminated in developing an efficient, tournament-style method for incorporating dynamic search spaces into ICDARTS. The final set of improvements to the search space consisted of new cell discretization protocols that expanded the space of possible cells that could be discovered by ICDARTS and allowed fairer comparisons between incoming edges. 

\begin{figure}[ht]
\centering
\resizebox{0.7\columnwidth}{!}{
\begin{tikzpicture}[->,>=stealth',auto,node distance=0.75cm,
  thick,main node/.style={circle,draw,font=\sffamily\Large\bfseries}]
\node[rectangle, rounded corners, minimum width=2.5cm, minimum height=0.5cm,text centered, draw=black, xshift=4.0cm, yshift=-4.00cm] (0) {Input};
\node[rectangle, rounded corners, minimum width=2.5cm, minimum height=0.5cm,text centered, draw=black, yshift=-0.25cm] (1) [below of=0, fill=green!20] {Normal Cell};
\node[rectangle, rounded corners, minimum width=2.5cm, minimum height=0.5cm,text centered, draw=black] (2) [below of=1, fill=green!20] {Normal Cell};
\node[rectangle, rounded corners, minimum width=2.5cm, minimum height=0.5cm,text centered, draw=black] (3) [below of=2, fill=yellow!20] {Reduction Cell};

\node[rectangle, rounded corners, minimum width=2.5cm, minimum height=0.5cm,text centered, draw=black, yshift=-0.25cm] (4) [below of=3, fill=green!20] {Normal Cell};
\node[rectangle, rounded corners, minimum width=2.5cm, minimum height=0.5cm,text centered, draw=black] (5) [below of=4, fill=green!20] {Normal Cell};
\node[rectangle, rounded corners, minimum width=2.5cm, minimum height=0.5cm,text centered, draw=black] (6) [below of=5, fill=yellow!20] {Reduction Cell};

\node[rectangle, rounded corners, minimum width=2.5cm, minimum height=0.5cm,text centered, draw=black, yshift=-0.25cm] (7) [below of=6, fill=green!20] {Normal Cell};
\node[rectangle, rounded corners, minimum width=2.5cm, minimum height=0.5cm,text centered, draw=black] (8) [below of=7, fill=green!20] {Normal Cell};

\node[rectangle, rounded corners, minimum width=2.5cm, minimum height=0.5cm,text centered, draw=black, yshift=-0.25cm] (9) [below of=8] {Output};

\node[rectangle, rounded corners, minimum width=2.5cm, minimum height=0.5cm,text centered, draw=black, yshift=-3cm] (10) [below of=9] {Joint Loss};

\node[rectangle, align=left, rounded corners, minimum width=3.00cm, minimum height=2.25cm, draw=black ] (11) [below of = 1] {};
\node[rectangle, align=left, rounded corners, minimum width=3.00cm, minimum height=2.25cm, draw=black ] (12) [below of = 4] {};
\node[rectangle, align=left, rounded corners, minimum width=3.00cm, minimum height=1.5cm, draw=black, yshift=0.38cm ] (13) [below of = 7] {};
\node[rectangle, align=left, rounded corners, minimum width=3.25cm, minimum height=6.75cm, draw=black, yshift=0.38cm ] (14) [below of = 4] {};

\draw [->] (0.south) to [right] (1.north);
\draw [-] (1.south) to [right] (2.north);
\draw [-] (2.south) to [right] (3.north);
\draw [->] (3.south) to [right] (4.north);
\draw [-] (4.south) to [right] (5.north);
\draw [-] (5.south) to [right] (6.north);
\draw [->] (6.south) to [right] (7.north);
\draw [-] (7.south) to [right] (8.north);
\draw [->] (8.south) to [right] (9.north);
\draw [->] (9.south) to [right] (10.north);

\node[rectangle, rounded corners, minimum width=2.5cm, minimum height=0.5cm,text centered, draw=black, xshift=8.0cm] (15) {Input};
\node[rectangle, rounded corners, minimum width=2.5cm, minimum height=0.5cm,text centered, draw=black, yshift=-0.25cm] (16) [below of=15, fill=green!20] {Normal Cell};
\node[rectangle, rounded corners, minimum width=2.5cm, minimum height=0.5cm,text centered, draw=black] (17) [below of=16, fill=green!20] {Normal Cell};
\node[rectangle, rounded corners, minimum width=2.5cm, minimum height=0.5cm,text centered, draw=black] (18) [below of=17, fill=green!20] {Normal Cell};
\node[rectangle, rounded corners, minimum width=2.5cm, minimum height=0.5cm,text centered, draw=black] (19) [below of=18, fill=green!20] {Normal Cell};
\node[rectangle, rounded corners, minimum width=2.5cm, minimum height=0.5cm,text centered, draw=black] (20) [below of=19, fill=green!20] {Normal Cell};
\node[rectangle, rounded corners, minimum width=2.5cm, minimum height=0.5cm,text centered, draw=black] (21) [below of=20, fill=green!20] {Normal Cell};
\node[rectangle, rounded corners, minimum width=2.5cm, minimum height=0.5cm,text centered, draw=black] (22) [below of=21, fill=yellow!20] {Reduction Cell};

\node[rectangle, rounded corners, minimum width=2.5cm, minimum height=0.5cm,text centered, draw=black, yshift=-0.25cm] (23) [below of=22, fill=green!20] {Normal Cell};
\node[rectangle, rounded corners, minimum width=2.5cm, minimum height=0.5cm,text centered, draw=black] (24) [below of=23, fill=green!20] {Normal Cell};
\node[rectangle, rounded corners, minimum width=2.5cm, minimum height=0.5cm,text centered, draw=black] (25) [below of=24, fill=green!20] {Normal Cell};
\node[rectangle, rounded corners, minimum width=2.5cm, minimum height=0.5cm,text centered, draw=black] (26) [below of=25, fill=green!20] {Normal Cell};
\node[rectangle, rounded corners, minimum width=2.5cm, minimum height=0.5cm,text centered, draw=black] (27) [below of=26, fill=green!20] {Normal Cell};
\node[rectangle, rounded corners, minimum width=2.5cm, minimum height=0.5cm,text centered, draw=black] (28) [below of=27, fill=green!20] {Normal Cell};
\node[rectangle, rounded corners, minimum width=2.5cm, minimum height=0.5cm,text centered, draw=black] (29) [below of=28, fill=yellow!20] {Reduction Cell};

\node[rectangle, rounded corners, minimum width=2.5cm, minimum height=0.5cm,text centered, draw=black, yshift=-0.25cm] (30) [below of=29, fill=green!20] {Normal Cell};
\node[rectangle, rounded corners, minimum width=2.5cm, minimum height=0.5cm,text centered, draw=black] (31) [below of=30, fill=green!20] {Normal Cell};
\node[rectangle, rounded corners, minimum width=2.5cm, minimum height=0.5cm,text centered, draw=black] (32) [below of=31, fill=green!20] {Normal Cell};
\node[rectangle, rounded corners, minimum width=2.5cm, minimum height=0.5cm,text centered, draw=black] (33) [below of=32, fill=green!20] {Normal Cell};
\node[rectangle, rounded corners, minimum width=2.5cm, minimum height=0.5cm,text centered, draw=black] (34) [below of=33, fill=green!20] {Normal Cell};
\node[rectangle, rounded corners, minimum width=2.5cm, minimum height=0.5cm,text centered, draw=black] (35) [below of=34, fill=green!20] {Normal Cell};
\node[rectangle, rounded corners, minimum width=2.5cm, minimum height=0.5cm,text centered, draw=black, yshift=-0.25cm] (36) [below of=35] {Output};

\node[rectangle, align=left, rounded corners, minimum width=3.00cm, minimum height=5.25cm, draw=black ] (37) [below of = 18] {};
\node[rectangle, align=left, rounded corners, minimum width=3.00cm, minimum height=5.25cm, draw=black ] (38) [below of = 25] {};
\node[rectangle, align=left, rounded corners, minimum width=3.00cm, minimum height=4.5cm, draw=black, yshift=0.38cm ] (39) [below of = 32] {};
\node[rectangle, align=left, rounded corners, minimum width=3.25cm, minimum height=15.75cm, draw=black, yshift=0.38cm] (40) [below of = 25] {};

\draw [->] (15.south) to [right] (16.north);
\draw [-] (16.south) to [right] (17.north);
\draw [-] (17.south) to [right] (18.north);
\draw [-] (18.south) to [right] (19.north);
\draw [-] (19.south) to [right] (20.north);
\draw [-] (20.south) to [right] (21.north);
\draw [-] (21.south) to [right] (22.north);
\draw [->] (22.south) to [right] (23.north);
\draw [-] (23.south) to [right] (24.north);
\draw [-] (24.south) to [right] (25.north);
\draw [-] (25.south) to [right] (26.north);
\draw [-] (26.south) to [right] (27.north);
\draw [-] (27.south) to [right] (28.north);
\draw [-] (28.south) to [right] (29.north);
\draw [->] (29.south) to [right] (30.north);
\draw [-] (30.south) to [right] (31.north);
\draw [-] (31.south) to [right] (32.north);
\draw [-] (32.south) to [right] (33.north);
\draw [-] (33.south) to [right] (34.north);
\draw [-] (34.south) to [right] (35.north);
\draw [->] (35.south) to [right] (36.north);

\node[rectangle, align=left, rounded corners, minimum width=3.25cm, minimum height=5.75cm, draw=black, yshift=-8.00cm] (41)  {};
\node[rectangle, align=left, rounded corners, minimum width=1.0cm, minimum height=0.75cm, draw=black, yshift=-6.0cm] (42)  {\(n_0\)};

\node[rectangle, align=left, rounded corners, minimum width=1.0cm, minimum height=0.75cm, draw=black, xshift=-0.75cm, yshift=-0.5cm] (45) [below of=42]{\(n_1\)};
\node[rectangle, align=left, rounded corners, minimum width=1.0cm, minimum height=0.75cm, draw=black, xshift=0.25cm, yshift=-2cm] (46) [below of=42]{\(n_2\)};
\node[rectangle, align=left, rounded corners, minimum width=1.0cm, minimum height=0.75cm, draw=black, yshift=-3.25cm] (47) [below of=42]{\(n_3\)};

\draw [red!60,->] ([xshift=-0.4 cm]42.south) to [out=-135,in=45] ([xshift=-0.1 cm]45.north);
\draw [green!60,->] ([xshift=-0.3 cm]42.south) to [out=-135,in=45] ([xshift=0.0 cm]45.north);
\draw [blue!60,->] ([xshift=-0.2 cm]42.south) to [out=-135,in=45] ([xshift=0.1 cm]45.north);

\draw [red!60,->] ([xshift=-0.1 cm]42.south) to [out=-70,in=105] ([xshift=0.1cm]46.north);
\draw [green!60,->] ([xshift=-0.0 cm]42.south) to [out=-70,in=105] ([xshift=0.2cm]46.north);
\draw [blue!60,->] ([xshift=0.1 cm]42.south) to [out=-70,in=105] ([xshift=0.3cm]46.north);

\draw [red!60,->] ([xshift=0.1 cm]45.south) to [out=-45,in=135] ([xshift=-0.3cm]46.north);
\draw [green!60,->] ([xshift=0.2 cm]45.south) to [out=-45,in=135] ([xshift=-0.2cm]46.north);
\draw [blue!60,->] ([xshift=0.3 cm]45.south) to [out=-45,in=135] ([xshift=-0.1cm]46.north);

\draw [blue!60,->] ([xshift=0.4 cm]42.south) to [out=-35,in=45] ([yshift=-0.1 cm]47.east);
\draw [green!60,->] ([xshift=0.3 cm]42.south) to [out=-35,in=45] ([yshift=0.0 cm]47.east);
\draw [red!60,->] ([xshift=0.2 cm]42.south) to [out=-35,in=45] ([yshift=0.1 cm]47.east);

\draw [red!60,->] ([xshift=-0.3 cm]45.south) to [out=-135,in=135] ([yshift=-0.1 cm]47.west);
\draw [green!60,->] ([xshift=-0.2 cm]45.south) to [out=-135,in=135] ([yshift=0.0 cm]47.west);
\draw [blue!60,->] ([xshift=-0.1 cm]45.south) to [out=-135,in=135] ([yshift=0.1 cm]47.west);

\draw [blue!60,->] ([xshift=0.1 cm]46.south) to [out=-135,in=45] ([xshift=0.1 cm]47.north);
\draw [green!60,->] ([xshift=0.0 cm]46.south) to [out=-135,in=45] ([xshift=0.0 cm]47.north);
\draw [red!60,->] ([xshift=-0.1 cm]46.south) to [out=-135,in=45] ([xshift=-0.1 cm]47.north);


\node[rectangle, align=left, rounded corners, minimum width=3.25cm, minimum height=5.75cm, draw=black, xshift=12.0cm, yshift=-8.0cm] (61)  {};
\node[rectangle, align=left, rounded corners, minimum width=1.0cm, minimum height=0.75cm, draw=black, xshift=12.0cm, yshift=-6.0cm] (62)  {\(n_0\)};

\node[rectangle, align=left, rounded corners, minimum width=1.0cm, minimum height=0.75cm, draw=black, xshift=-0.75cm, yshift=-0.5cm] (65) [below of=62]{\(n_1\)};
\node[rectangle, align=left, rounded corners, minimum width=1.0cm, minimum height=0.75cm, draw=black, xshift=0.25cm, yshift=-2cm] (66) [below of=62]{\(n_2\)};
\node[rectangle, align=left, rounded corners, minimum width=1.0cm, minimum height=0.75cm, draw=black, yshift=-3.25cm] (67) [below of=62]{\(n_3\)};

\draw [green,->] ([xshift=-0.2 cm]62.south) to [out=-135,in=45] ([xshift=0.0 cm]65.north);

\draw [red,->] ([xshift=0.1 cm]62.south) to [out=-70,in=105] ([xshift=0.3cm]66.north);

\draw [red,->] ([xshift=0.1 cm]65.south) to [out=-45,in=135] ([xshift=-0.3cm]66.north);

\draw [blue,->] ([xshift=0.4 cm]62.south) to [out=-35,in=45] ([yshift=-0.1 cm]67.east);

\draw [blue,->] ([xshift=-0.1 cm]65.south) to [out=-135,in=135] ([yshift=0.1 cm]67.west);

\draw [green,->] ([xshift=0.0 cm]66.south) to [out=-135,in=45] ([xshift=0.0 cm]67.north);

\node[align=center, minimum width=2.5cm, minimum height=1.75cm, yshift=0.3 cm] (60) [above of = 0] {Search \\Network};
\node[align=center, minimum width=2.5cm, minimum height=1.75cm, yshift=0.3 cm] (61) [above of = 15] {Evaluation\\ Network};
\node[align=center, minimum width=2.5cm, minimum height=1.75cm, yshift=2.75cm] (62) [above of = 41] {Continuous\\ Cell};
\node[align=center, minimum width=2.5cm, minimum height=1.75cm, xshift=4cm, yshift=-6.25cm] (63) [above of = 61] {Discetized\\ Cell};

\draw [->, draw=red] (60.north) to [bend left] node [red, above, align=center, xshift=-0.5cm, yshift=0.5cm]  {Discretize\\ Network} (61.west);

\draw [->] (36.west) -| (10.south);

\end{tikzpicture}}
\caption{Overview of the CDARTS NAS algorithm, including search (left) and evaluation (right) networks with examples of continuous and discretized cells. }
\label{cdarts-fig}
\end{figure}
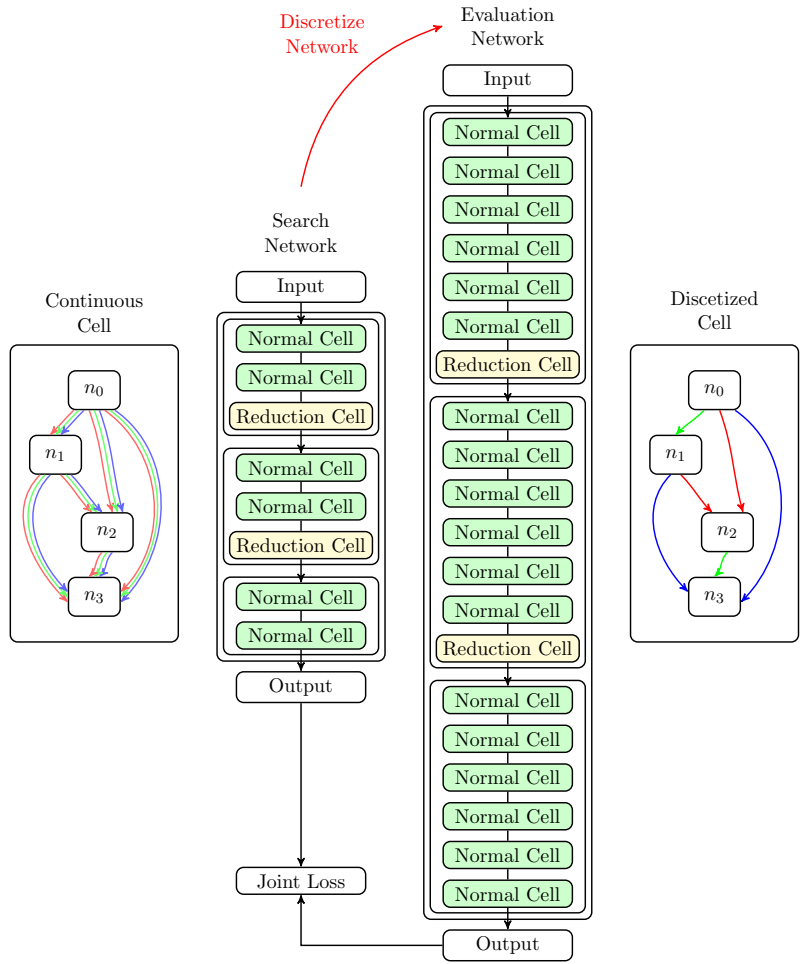

\section{Related Work}
Early automatic neural architecture search algorithms achieved remarkable performance on various image processing benchmarks \citep{elsken2019neural}. These methods' search strategies relied on various techniques, including reinforcement learning \citep{zoph2017neural, zoph2018learning} and evolutionary algorithms \citep{xie2017genetic,real2019regularized,elsken2019efficient,yang2020cars} to discover optimal network architectures given a large search space. Although these algorithms perform well, they involve evaluating many candidate architectures and can be computationally expensive. Weight sharing \textit{one-shot} methods have been proposed as one solution to this issue \citep{brock2017smash,li2019random, pham2018efficient}. These methods involve training one model, typically represented as one over-parameterized network graph, and iteratively sampling child models, or paths within the network graph, that share weights throughout the search process. Following training, the optimal model is selected by ranking the performance of the child models \citep{cai2020onceforall,guo2020single}. By harnessing the weight-sharing mechanism, \textit{one-shot} family models can evaluate high-capacity architectures within a few GPU days. The DARTS algorithm introduced a new variant of the weight-sharing model. Rather than searching over discrete architectures within the main search graph, DARTS searches over a continuous space of architectures by using stochastic gradient descent to optimize a mixture of weights corresponding to each edge of the main search graph. Then, in a separate retraining phase, the continuous architectures are "discretized" by selecting the set of \textit{k} edges at each node corresponding to the weights with the highest probabilities. By applying this approach, DARTS finds high-quality architectures with reduced computational cost \citep{liu2019darts, https://doi.org/10.48550/arxiv.1808.05377}.

The DARTS algorithm has its shortcomings. One is that the different search and retraining phases may result in two independent networks with limited correlation \citep{cai2018proxylessnas, yu2020cyclic}. One follow-up publication \citep{xie2020snas}, argues that the inconsistency between the search and evaluation networks results from removing operations between the search and evaluation networks that adds bias to the loss function and results in the need to retrain the network. The authors' solution is Stochastic Neural Architecture Search (SNAS), a framework that trains network and architecture weights on the same round of backpropagation. As a part of this process, one-hot random variables are used as masks to select operations in the search graph. The resulting method has improved stability and efficiency. The ProxylessNAS algorithm \citep{cai2018proxylessnas} was also designed to address this discrepancy by searching for a final, deep target network instead of a shallow, intermediate (or "proxy") network. This model minimizes computational expense by binarizing its architecture parameters so that only a single path is active at a time. Another approach, Progressive DARTS (P-DARTS) \citep{chen2019progressive,chen2021progressive}, addresses the disparity between the networks of the search and retraining phases by progressively increasing the depth of the search network until it reaches that of the evaluation network. In addition, this method further avoids high computational expense by gradually dropping lower-scoring candidate operations as the search process progresses. 

Cyclic DARTS (CDARTS)~\citep{yu2020cyclic} addresses the discrepancy between the over-parameterized search network and the discretized retraining network by introducing a cyclic feedback mechanism into the search phase. In each iteration of its search phase, its search network and an intermediate evaluation network are optimized using a joint learning process. The evaluation network is generated at the beginning of each search iteration by discretizing the search network given its connection weights at that stage in the search process. By providing feedback to the search network cyclically, the evaluation network serves as an effective proxy for the final network architecture.

Our approach introduces critical improvements to the original CDARTS algorithm, which further minimizes discrepancies between the search and evaluation networks by modifying the loss functions of each. 
We first correct the inconsistency between the evaluation network loss functions in the search and retraining phases by eliminating the feature distillation loss term, which measures the distance between the logits of the search and evaluation networks, from the search phase loss function. This change removes the dependency of the search phase evaluation network's weights on the search network's loss and results in an evaluation network loss function that is consistent across the search and retraining phases. 

Next, we modified CDARTS to optimize its search and evaluation network weights on the same dataset. In the original version of CDARTS, these weights are trained on separate splits of the training dataset, which introduces potential bias in the joint learning phase. 

Finally, we incorporate the feature distillation loss term into the search network's loss function so that this network's weights are updated based on feedback supplied by the evaluation network rather than the other way around. 
The culmination of these changes is an updated version of the CDARTS algorithm that improves upon the stability of the original algorithm by incorporating modifications that ensure the search and evaluation networks more closely resemble each other. 

\section{Methods}
This section details our improvements to the CDARTS algorithm. We first describe the original DARTS and CDARTS architectures, from the directed acyclic graph structures representing their cell motifs to the network templates that dictate how they are arranged within the search and evaluation networks. We then summarize the CDARTS algorithm's objective and loss functions. 
The remainder of this section introduces improvements to the CDARTS algorithm's joint optimization process that result in processes for training the search and evaluation networks that more closely resemble each other. We propose alternatives to the \(zero\) layer option from the DARTS (and CDARTS) search space alongside these algorithmic changes. Then, we propose ablation studies on the ICDARTS algorithm and network structure. Afterward, we propose methods for expanding the search space of ICDARTS, including exploring alternate operation spaces and higher capacity cell discretization approaches. With these methods, we propose a novel algorithm for incorporating a dynamic search space into ICDARTS.

\subsection{CDARTS}
In this subsection, we will review the CDARTS algorithm as presented in \citep{xu2020pcdarts}.
\subsubsection{CDARTS Architecture}
 Figure \ref{cdarts-fig} gives an overview of the CDARTS algorithm, illustrates the search and evaluation network templates, and shows examples of continuous and discretized cell structures. In both DARTS and CDARTS, the architectures of the search and evaluation networks are composed of cell motifs represented by directed acyclic graphs (DAG), which serve as the building blocks of both networks. As discussed in \citep{liu2018progressive} and \citep{liu2019darts}, each cell graph consists of \(N\) nodes, each of which denote some feature representation. Each directed edge \(i,j\) within the graph represents a particular operation, $o^{(i,j)}$, that is applied to a node, \(x_i\), to produce \(x_j\). The operation space of DARTS consists of operations with (e.g.  convolution) and without (e.g. maximum and average pooling operations, skip connections, and \(zero\) operations) learned weights. 
\color{black}
DARTS achieves a continuous search space by weighting the candidate operations at each edge by their corresponding \(\alpha\) values:

\begin{equation}
    \begin{aligned}
\bar{o}^{(i,j)}(x_i) = \sum_{o\in O} \frac{\exp(\alpha_o^{(i,j)})}{\sum_{o' \in O} \exp(\alpha_{o'}^{(i,j)})} o(x_i)
    \end{aligned}
\end{equation}

These \(\alpha^{(i,j)}\) values consist of vectors of size $\lvert O \rvert$ that parameterize the candidate operation strength for each edge in the graph. Each cell accepts inputs from two previous cells, \(c_{k-2}\) and \(c_{k-1}\). Each node's output is computed by taking the weighted sum of the outputs of the preceding nodes and outputs of the two previous cells: \(x_j = \sum_{}\hat{o}^{(i,j)(x_i)}\). The outputs of each node in a cell are concatenated to produce the final output of the cell.
Two types of cells are optimized during the search phase: normal and reduction. The normal cells produce outputs with the same spatial dimensions as their inputs. In contrast, the reduction cells produce outputs with spatial dimensions that have been reduced by a factor of two by applying a stride of 2 two in its operations \citep{zoph2018learning}.

To construct the evaluation networks, the continuous search cells are discretized by removing all but the edges with top \(k=2\) alpha values for each node, such that \(o^{(i,j)} = \argmax_{o \in O}\alpha_o^{(i,j)}\). These cells are then stacked to form a deeper network specified by a template (again, see Figure \ref{cdarts-fig})  \citep{zoph2018learning, real2019regularized, liu2018progressive}. 

In addition to the network structure pictured in Figure \ref{cdarts-fig}, both networks include an auxiliary head structure, which combines the output of the final normal cell with that of both reduction cells to get the network's final output \citep{yu2020cyclic}.

\subsubsection{CDARTS Algorithm}
As discussed in \citep{liu2019darts}, the DARTS algorithm searches optimal cell structures for the full evaluation network by using stochastic gradient descent to optimize connection weights within the continuous cells that form the search network. 
 The objective of CDARTS' search phase is to identify the connection weights, \(\alpha\), and search network weights, \(w_s\) that satisfy the following bilevel optimization problem: 
 
\begin{equation}
    \begin{aligned}
    \min_\alpha L_{val}(w^*_s, \alpha) \\  
    s.t. \;  w^*_s(\alpha) = \argmin_{w_s} L_{train}(w_s, \alpha)
    \end{aligned}
\end{equation}

The connection weights, \(\alpha\), are optimized given a validation dataset, but the search network weights, \(w_s\), are learned given a separate training dataset. Afterward, a retraining phase takes place, in which a deeper evaluation network is generated by discretizing the cells discovered in the search phase. The network is retrained from scratch  and evaluated on a new test dataset \citep{liu2019darts, yu2020cyclic}  

CDARTS expands upon DARTS by introducing a cyclic feedback mechanism between the search network and an intermediate version of its evaluation network during its search phase. The intermediate evaluation network is generated at the beginning of each iteration of the algorithm and is intended to resemble the final evaluation network of the retraining phase. This innovation allows for the search and evaluation networks to be optimized jointly and enables them to better mirror each other in terms of performance and learned features. The following equation represents the joint optimization process, given the search and intermediate evaluation network weights, \(w_S\) and \(w_E\) and connection weights, \(\alpha\):

\begin{equation}
    \begin{aligned}
\min_\alpha L_{val}(w_E^*, w_S^*, \alpha) \\ s.t.~~~ w^*_S = \argmin_{w_S} L_{train}(w_S, \alpha)\\ w^*_E = \argmin_{w_E} L_{train}(w_E, \alpha)
    \end{aligned}
\end{equation}

CDARTS consists of two phases: \textit{pre-training} and \textit{joint learning}, which are repeated cyclically until convergence. In the \textit{pre-training} phase, the weights of the search and intermediate evaluation networks are trained for a limited number of epochs. The search network weights \(w_S\) are optimized according to the following equation:

\begin{equation}
    w^*_S = \argmin_{w_S} L_{train}^S(w_S, \alpha)
\end{equation}

where \(L_{train}^S\) denotes the cross entropy loss function. The intermediate evaluation network is then generated given the learned \(\alpha\) weights by following the same discretization procedure used in DARTS. The intermediate evaluation network weights \(w_E\) are finally optimized on the validation set according to the loss function:

\begin{equation}
    w^*_E = \argmin_{w_E} L_{val}^E(w_E, \bar{\alpha})
\end{equation}

where \(\bar{\alpha}\) represents the discretized cell architetures resulting from \(\alpha\).

During the \textit{joint training} phase, the \(\alpha\) and \(w_E\) weights are updated based on the cross entropy losses of the search and evaluation networks and a soft-target cross-entropy loss, which measures the distance between the logits of the search and evaluation networks. The following equation represents this joint optimization task:

\begin{equation}
    \label{eq:alpha_we_loss}
    \begin{aligned}
    \alpha^*, w^*_E = \argmin_{\alpha, w_E} L_{val}^S(w_S^*, \alpha) + L_{val}^E(w_E, \bar{\alpha}) \\ + \lambda L_{val}^{S,E}(w_S^*, \alpha, w_E, \bar{\alpha})
    \end{aligned}
\end{equation}

By minimizing the \(L_{val}^S(w_S^*, \alpha)\) term, the \(\alpha\) parameter is optimized given the fixed search network weights \(w_S^*\). The \(L_{val}^E(w_E, \bar{\alpha})\) variable serves to optimize the evaluation network weights given \(\bar{\alpha}\) as a fixed parameter. Finally, the soft-target cross-entropy loss term, \(L_{val}^{S,E}(w_S^*, \alpha, w_E, \bar{\alpha})\) allows for the transfer of knowledge from the evaluation network to the search network by applying the features learned from the evaluation network to the \(\alpha\) parameter of the search network. This term is formulated as:

\begin{equation}
\label{eq:soft-target-ce}
    L_{val}^{S,E}(w_S^*, \alpha, w_E, \bar{\alpha}) = \frac{T^2}{N} \sum_{i=1}^N (p(w_E, \bar{\alpha})log(\frac{p(w_E, \bar{\alpha})}{q(w_S^*, \alpha)}))
\end{equation}

where \(N\) is the number of training samples, \(T\) is the temperature coefficient, and \(p\) and \(q\) are the feature logits of the search and evaluation networks, respectively.  \(p\) and \(q\) are calculated given the features of the search and evaluation networks \(f_i^S\) and \(f_i^E\):

\begin{equation}
    \begin{aligned}
    p(w_E, \bar{\alpha}) = \frac{\exp(f_i^E/T)}{\sum_j\exp(f_j^E/T)} ,\\
    q(w_S^*, \alpha) = \frac{\exp(f_i^S/T)}{\sum_j\exp(f_j^S/T)} 
    \end{aligned}
\end{equation}

The result of this \textit{joint training} phase is sufficient knowledge transfer between the search and evaluation networks that ensures both perform similarly and learn similar features \citep{yu2020cyclic}.

\begin{algorithm}[t]
\caption{ICDARTS Search Phase}
\label{alg:icdarts}
\begin{algorithmic}
   \State {\bfseries Input:} Datasets $train$ and $val$, search and evaluation iterations $S_S$ and $S_E$, update iterations $S_U$, architecture hyperparameter $\alpha$, and weights $w_S$ and $w_E$ for search $S$ and evaluation $E$ networks
   \State {\bfseries Output:}  Evaluation network $E$
   \State Initialize $\alpha$ randomly
   \State Initialize $w_S$
   \For{each search step $i \in [0,S_S]$} 
   \If{$i Mod S_U$} 
   \State Discretize $\alpha$ to $\bar{\alpha}$ by selecting the top $k$
   \State Generate $E$ with $\bar{\alpha}$ 
   \For{each evaluation step $j \in [0, S_E] $} 
   \State{Calculate $L^E_{val}$}
   \State Update $w_E$ according to Eq. 5
   \EndFor
   \EndIf
   \State Calculate $L^S_{val}$, $L^E_{val}$, and $L^{S,E}_{val}$
   \State Update $\alpha$ according to Eq. 9
   \vspace{0.5mm}
   \State Calculate $L^S_{train}$ and $L^{S,E}_{train}$
   \State Update $w_S$ according to Eq. 10
   \State Calculate $L^E_{train}$
   \State Update $w_E$ according to Eq. 11
   \EndFor
\end{algorithmic}
\end{algorithm}

\subsection{ICDARTS}

\subsubsection{Algorithm Updates}

\begin{table}[t]
\caption{ICDARTS 'Zero' Operation Return Type Configurations}
\begin{center}
\begin{tabular}{|c|c|c|c|}
\hline
\textbf{Zero Return}&\multicolumn{3}{|c|}{\textbf{Zero Return Type}} \\
\cline{2-4} 
\textbf{Configuration} & \textbf{\textit{Search}}& \textbf{\textit{Evaluation}}& \textbf{\textit{Retraining}} \\
\hline
V0  & zero & not included & not included \\
\hline
V1  & not included & not included & not included \\
\hline
V2  & random & random & random \\
\hline
V3  & random & random & zero \\
\hline
V4  & random & zero & zero \\
\hline
\end{tabular}
\label{tab-zero-configs}
\end{center}
\label{tab_zero_types}
\end{table}

CDARTS strives to ensure that the continuous search cells optimized within its search network effectively translate to discretized cells within a deeper evaluation network that performs well and learns similar features. However, when reviewing the CDARTS algorithm and its loss functions, we identified multiple issues that limited correlation between the networks optimized in the search and evaluation phases and introduced changes that address each. 

First, we noted that the loss function used for updating the intermediate evaluation network during the search phase (see Equation \ref{eq:alpha_we_loss}) was formulated so that the network's weights are dependent on both the search network's loss and the soft-target cross-entropy loss. This was in contrast to the loss function used for training the evaluation network during the retraining phase, which depended only on the evaluation network's loss. In order to remove the dependence of the search phase evaluation network on the additional terms, we modify CDARTS' joint optimization approach so that the weights of the evaluation network are no longer dependent on the loss of the search network but only that of the evaluation network.  

Next, we shift the soft-target cross-entropy loss term, \(\lambda L_{val}^{S,E}(w_S, \alpha, w_{E}, \bar{\alpha})\), to the equation for updating the search network weights. This change allows this term to be retained in the joint learning phase and better serves the purpose of facilitating the transfer of knowledge from the evaluation network's weights to those of the search network. 

Finally, we noted that CDARTS optimizes the search and evaluation networks' weights on two separate datasets. 
This approach injects unnecessary bias into the joint learning phase, particularly in the case of the soft-target cross-entropy loss term, where CDARTS is attempting to get the two networks to produce the same output even though they are trained on separate datasets.
With this in mind, we modify the function for updating the evaluation network weights so that it is dependent on the evaluation network's loss on the training rather than the validation dataset. 

The loss functions for training the \(\alpha\), \(w_S\), and \(w_E\) weights are now:
\begin{equation}
    \label{eq:alpha_we_loss_mod}
    \alpha^* = \argmin_{\alpha} L_{val}^S(w_S^*, \alpha) + \lambda L_{val}^{S,E}(w_S^*, \alpha, w_E^*, \bar{\alpha})
\end{equation}

\begin{equation}
    w^*_S = \argmin_{w_S} L_{train}^S(w_S, \alpha) + \lambda L_{train}^{S,E}(w_S, \alpha, w_{E}, \bar{\alpha})
\end{equation}

\begin{equation}
    w^*_E = \argmin_{w_E} L_{train}^{E}(w_{E}, \bar{\alpha})
\end{equation}

Algorithm \ref{alg:icdarts} presents the algorithm for the ICDARTS search phase and shows how it incorporates our reformulations to the CDARTS loss functions as first introduced in our conference paper \citep{herron2022icdarts}. The result of our changes is a search phase evaluation network that is a better proxy for that of the retraining phase. The search phase evaluation network no longer depends on a loss term that will be unavailable during retraining. Additionally, the soft target cross-entropy loss term is now used for updating the search network's weights so that feedback from the evaluation network is supplied to the search network, which better aligns with the original intended purpose of this term. Finally, training the networks' weights on the same dataset ensures that the networks perform similarly rather than forcing them to produce the same output given two different datasets.

\begin{table}[h]
\caption{ICDARTS Template Ablations and Replacements}
\centering
\begin{center}
\resizebox{0.6\columnwidth}{!}{%
\begin{tabular}{cc}
\hline
\multicolumn{1}{|c|}{\textbf{Ablation}}                & \multicolumn{1}{c|}{\textbf{Replacement}}          \\ \hline
\multicolumn{1}{|c|}{Pooling Operations}               & \multicolumn{1}{c|}{-}                             \\ \hline
\multicolumn{1}{|c|}{Identity Operation}               & \multicolumn{1}{c|}{-}                             \\ \hline
\multicolumn{1}{|c|}{Dilated Convolution Operations}   & \multicolumn{1}{c|}{-}                             \\ \hline
\multicolumn{1}{|c|}{Seperable Convolution Operations} & \multicolumn{1}{c|}{-}                             \\ \hline
\multicolumn{1}{|c|}{Auxilary Heads}                   & \multicolumn{1}{c|}{-}                             \\ \hline
\multicolumn{1}{|c|}{\multirow{3}{*}{Stemming Layer}}  & \multicolumn{1}{c|}{Identity}                      \\ \cline{2-2} 
\multicolumn{1}{|c|}{}                                 & \multicolumn{1}{c|}{Convolution (k=1), Batch Norm} \\ \cline{2-2} 
\multicolumn{1}{|c|}{}                                 & \multicolumn{1}{c|}{Concatenated Inputs}            \\ \hline
\multicolumn{1}{|c|}{\multirow{3}{*}{Reduce Cells}}    & \multicolumn{1}{c|}{Average Pooling}               \\ \cline{2-2} 
\multicolumn{1}{|c|}{}                                 & \multicolumn{1}{c|}{Max Pooling}                   \\ \cline{2-2} 
\multicolumn{1}{|c|}{}                                 & \multicolumn{1}{c|}{Convolution (k=1, stride=2)}   \\ \hline                           
\end{tabular}
}
\label{tab_temp_ablation}
\end{center}
\end{table}
\begin{table}[h]
\caption{ICDARTS Algorithmic Ablation Study Routes with Loss Function Updates}
\centering
\begin{center}
\resizebox{0.6\columnwidth}{!}{%
\begin{tabular}{|p{2cm}|p{3.5cm}|p{3.5cm}|}
\hline
\multirow{2}{*}{\textbf{Algorithm}}&\multicolumn{2}{|p{7cm}|}{\centering{\textbf{Loss Function}}} \\
\cline{2-3}
 & \textbf{Route A} & \textbf{Route B} \\
\hline
CDARTS &  \multicolumn{2}{|c|}{\shortstack{\(\alpha \rightarrow L_{val} + L_{val}(J)\) \\ \(w_e \rightarrow  L_{val} + L_{val}(J)\) \\ \(w_s \rightarrow L_{trn}\)}}  \\
\hline
Ablation 1 
&  \shortstack{\(\alpha \rightarrow \) No Change 
\\ \(w_e \rightarrow\) No Change 
\\ \(w_s \rightarrow L_{trn} + L_{trn}(J)\) } 

& \shortstack{ \(\alpha \rightarrow \) No Change 
\\ \(w_e \rightarrow L_{trn} + L_{trn}(J)\) 
\\ \(w_s \rightarrow\) No Change }  \\

\hline
Ablation 2 & \shortstack{\(\alpha \rightarrow \) No Change \\ \(w_e \rightarrow  L_{val} \)\\ \(w_s \rightarrow\) No Change} & \shortstack{\(\alpha \rightarrow \) No Change \\ \(w_e \rightarrow  L_{trn} \)\\ \(w_s \rightarrow\) No Change } \\
\hline
ICDARTS &  \shortstack{\(\alpha \rightarrow\) No Change \\ \(w_e \rightarrow L_{trn}(J)\)\\ \(w_s \rightarrow \) No Change} & \shortstack{\(\alpha \rightarrow\) No Change\\ \(w_e \rightarrow \) No Change\\ \(w_s \rightarrow L_{trn} + L_{trn}(J)\)} \\
\hline
\end{tabular}
}
\label{tab-alg-ablation-routes}
\end{center}
\end{table}

\begin{table}[h]
\caption{ICDARTS Operation Search Spaces }
\label{tab:table-icdarts-ss}
\begin{center}
\resizebox{\columnwidth}{!}{
\begin{tabular}[b]{|c|c|c|c|c|}
\hline
Search Space & Category & Description & Operation & Parameters \\ \hline
\multirow{5}{0.75cm}{1} &  \multirow{5}{2.5cm}{Basic Operations} &  \multicolumn{1}{c|}{\multirow{5}{5cm}{Basic building blocks of more complex operations.}} & Convolution & \(k={3,5}\) \\ \cline{4-5}
 & &  \multicolumn{1}{c|}{} & Depthwise Convolution & \(k={3,5}\) \\ \cline{4-5}
 & &  \multicolumn{1}{c|}{} & ReLU & \\ \cline{4-5}
 & &  \multicolumn{1}{c|}{} & Leaky ReLU & \\ \cline{4-5}
 & &  \multicolumn{1}{c|}{} & BatchNorm & \\ \hline
 
\multirow{3}{0.75cm}{2} & \multirow{3}{2.5cm}{Simple Operations} 
 &  \multicolumn{1}{c|}{\multirow{3}{5cm}{Operations from early NAS literature including NASNet \cite{NASNet}.}} &  Minimum Convolution & \(k={3,5}\) \\ \cline{4-5}
 & &  \multicolumn{1}{c|}{} &  Standard Convolution & \(k={3,5}\) \\ \cline{4-5}
 & &  \multicolumn{1}{c|}{} &  Factorized Convolution & \(k={7,9}\) \\ \hline
 
\multirow{5}{0.75cm}{3} & \multirow{5}{2.5cm}{ICDARTS Search Space} 
 & \multicolumn{1}{c|}{\multirow{5}{5cm}{Widely-adopted search space used by DARTS and its derivatives.}} &  Identity* &  \\ \cline{4-5}
 & &  \multicolumn{1}{c|}{} &  Max Pooling* & \(k={3}\) \\ \cline{4-5}
 & &  \multicolumn{1}{c|}{} &  Average Pooling* & \(k={3}\) \\ \cline{4-5}
 & &  \multicolumn{1}{c|}{} &  Separable Convolution & \(k={3,5}\) \\ \cline{4-5}
 & &  \multicolumn{1}{c|}{} &  Dilated Convolution & \(k={3,5}\) \\ \hline

\multirow{3}{0.75cm}{4} & \multirow{3}{2.5cm}{MBConv Blocks} &  \multicolumn{1}{c|}{\multirow{3}{5cm}{State-of-the-art mobile convolution blocks.}} &  MBConv \cite{MobileNetV2} & \(k={3}\) \\ \cline{4-5}
 & &  \multicolumn{1}{c|}{} &  MBConv \cite{EfficientNet}  & \(k={3}\);  \(g={1}\)\\ \cline{4-5}
 & & \multicolumn{1}{c|}{} &  Fused-MBConv \cite{tan2021efficientnetv2} & \(k={3}\);  \(g={1}\)\\ \hline
 
\multicolumn{5}{l}{Note: Operations marked with * are included in all search spaces.} \\

\end{tabular}
\label{tab-search-spaces}
}
\end{center}
\end{table}

\begin{table}[h]
\caption{Combined Search Space used by Dynamic Search Algorithm}
\begin{center}
\begin{tabular}{|c|c|c|c|}
\hline
\textbf{Operation}     & \textbf{Parameters} & \textbf{Operation}     & \textbf{Parameters}\\ \hline
ReLU                   & -                   &
Leaky ReLU             & -                   \\ \hline
Sigmoid                & -                   &
Tanh                   & -                   \\ \hline
Batch Norm             & -                   &
Identity              & -                   \\ \hline
Convolution            & \(k={3,5}\)            &
Max Pooling           & \(k={3,5}\)            \\ \hline
Avg Pooling           & \(k={3,5}\)            &
Minimum Convolution    & \(k={3,5}\)            \\ \hline
Standard Convolution   & \(k={3,5}\)            &
Factorized Convolution & \(k={7,9}\)           \\ \hline
Separable Convolution  & \(k={3,5}\)            &
Dilated Convolution    & \(k={3,5}\)            \\ \hline
MBConv                 & \(k={3,5}\)               &
MBConvV2               & \(k={3,5}\) \(g={1,4,6}\) \\ \hline
Fused MBConv           & \(k={3}\) \(g={1,4,6}\)  & &\\ \hline
\end{tabular}
\end{center}
\label{tab-dynamic-search}
\end{table}
\begin{algorithm}[h]
\caption{ICDARTS Dynamic Search Space Algorithm}
\begin{algorithmic}
   \State {\bfseries Input:} Number of Tiers $T$, Pool of Layer Operations $O$, Maximum Operations per Set of Edges $O_{max}$
   \State {\bfseries Output:}  Optimal Operations, $o$ for Each Set of Edges
   \For{each $t \in {T,...1}$}
       \For{each $r \in {1,...,2t}$}
            \If{$t=3$}
                \State Randomly select $O_{max}$ layer options $o_{t,r}$ for each set of edges given $O$
            \Else
                \State Set $o_{t,r}$ to $o_{t-1, 2r} + o_{t-1, 2r+1}$ for each set of edges
            \EndIf
            
            \State Run ICDARTS Search Phase to optimize $o_{t,r}$ for each set of edges based on learned $\alpha$ values
            
            \If{$t=1$}
                \State Return $o_{t,r}$ for each set of edges
            \Else
                \State Update $o_{t,r}$ to top $50\%$ of $o_{t,r}$ by $\alpha$ value for each set of edges
            \EndIf
       \EndFor
   \EndFor
   
\end{algorithmic}
\label{alg-dynamic-search}
\end{algorithm}

\begin{figure*}[t]
\begin{center}
\includegraphics[width=0.9\columnwidth]{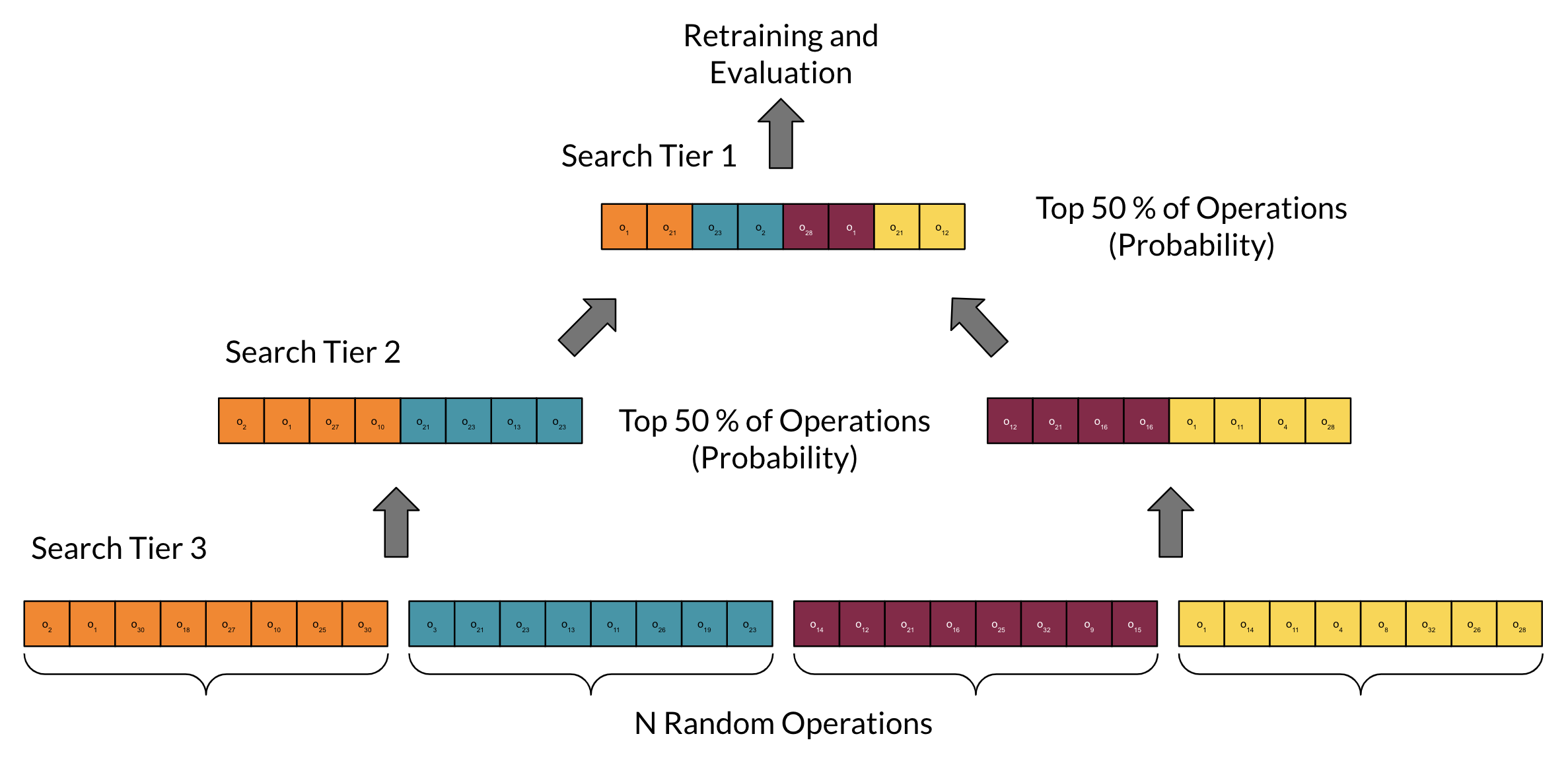}
\\
\caption{Tournament-Style Dynamic Search Space Algorithm for ICDARTS Overview with Tiers}
\label{fig-dynamic_search}
\end{center}
\end{figure*}

\subsubsection{Enabling None Layers in Discretized Networks}
The original DARTS method \citep{liu2019darts} includes \textit{zero} as a layer choice candidate operation. This option allows the NAS method to choose a layer that will produce no output. However, in DARTS and its derivatives, \citep{chen2019progressive, yu2020cyclic}, the \textit{zero} layer choice is not allowed when the network candidate operations are discretized as this option is constant and has a gradient of zero. Thus, its corresponding $\alpha$ weights cannot properly ascertain the importance of this operation. In practice, a large \(\alpha\) value and small \(\alpha\) value would produce the same output for the \textit{zero} layer.
Since the corresponding \(\alpha\) value is unlearnable, it is unclear why this operation choice is included during the search phase. In the following section, experimental results demonstrate comparable performance without including this operation during the search phase. Also explored is an alternative formulation of this operation that outputs randomly generated activations during search and is replaced with a no-op when the evaluation network is generated. The reformulation allows the alpha weight corresponding to the \textit{zero} operation to be learned without contributing to extracting features from previous layers. This approach assumes that if a layer contributes less than random noise, it should not be included in the final evaluation network.

To access the effectiveness of ICDARTS both with and without the \textit{zero} operation choice, we carried out five different experiments, each using a different configuration of \(zero\) operation return types. These configurations are all listed in Table \ref{tab_zero_types}. The V0 configuration is the same as the one used in the original CDARTS paper, in which the \(zero\) option is only included in the search network but ignored at discretization so that the deep evaluation networks and the final network used in retraining have no \textit{zero} layers. However, configurations \(V1-V4\) use the same number of layer options for the continuous search network and the discrete evaluation network. The traditional \(zero\) option returns a tensor of zeroes the size and shape of the input, and the \(random\) option similarly returns a tensor of uniform random values for each input. In addition to the experiments run using the CDARTS and ICDARTS search procedures, a separate set of experiments is run in which network cell motifs are randomly initialized eight times for each \(zero\) operation return protocol. These experiments aim to provide baseline performance results to contrast those of architectures discovered by both algorithms.

\subsection{Ablation Studies}

After making our initial set of improvements to the CDARTS algorithm, we conducted two ablation studies on the ICDARTS search space and algorithm. The first was conducted on the search space template of ICDARTS, including its operation choices, auxiliary heads, stemming layers, and reduce cells. Table \ref{tab_temp_ablation} contains a comprehensive listing of each ablation and any operations used to replace the ablation. Note that the none operation is left out of the set of operation choices by default.

The second ablation study was of the algorithmic changes to the CDARTS algorithm that resulted in ICDARTS. We considered two routes of incremental improvements to the original algorithm. The details of both are listed in Table \ref{tab-alg_ablation}.

\subsection{Alternate Search Spaces}

Upon completing the ablation study, we explored additional methods for expanding the limited search space of ICDARTS. This process involved exploring alternative sets of operation choices with varying complexities and cell discretization approaches that expanded the space of architectures that could be discovered during the search phase. We developed a novel tournament-based approach for incorporating dynamic search spaces in ICDARTS in conjunction with these experiments.

\subsubsection{Alternate Operation Spaces}

We first explored alternative operation search spaces for ICDARTS. We began by curating three additional operation search spaces for ICDARTS. Each search space was comprised of operations of varying complexities, ranging from the most basic operations to mobile convolution blocks from current state-of-the-art vision models. Each operation space and a brief description are listed in Table \ref{tab:table-icdarts-ss}. Note that pooling and identity operations, marked by the *, were included in each of the four search spaces, as they are all simple operations, yet have been included alongside more complex operations, such as in the widely adopted search space 3. Also, note that the minimum convolution operation is defined by a convolution followed by a batch norm, and the standard convolution operation consists of a ReLU activation function followed by a convolution and batch norm.

After running ICDARTS with operation search space, we combined the operations of all four search spaces and additional operations left out of the previous search spaces due to memory constraints into one master search space (see Table \ref{tab-dynamic-search}). In order to ensure that ICDARTS could traverse this large search space efficiently, we implemented an algorithm for incorporating dynamic search spaces into ICDARTS. Previous dynamic search algorithms partially inspired this novel, tournament-style dynamic search algorithm, including \citep{Dynamic_Search_Space} and \citep{SqueezeNAS}. As shown in Algorithm \ref{alg-dynamic-search} and Figure \ref{fig-dynamic_search}, our algorithm begins by randomly selecting \(O_{max}\) operations from the set of all operations, \(O\), for each set of cell edges. After running the search phase of ICDARTS to optimize the alpha weights corresponding to each edge, the operations corresponding to the top 50\% of alpha values are retained. The algorithm then proceeds to the next tier, forming its per-node operation set by combining the top operations from 2 runs in the previous tier. The process repeats until it reaches the topmost tier, after which it yields the final evaluation network to be evaluated in the retraining phase. 

\subsubsection{Alternate Search Cells}

\begin{figure*}[t]
\begin{center}
\begin{subfigure}{0.3\columnwidth}
\includegraphics[width=\columnwidth]{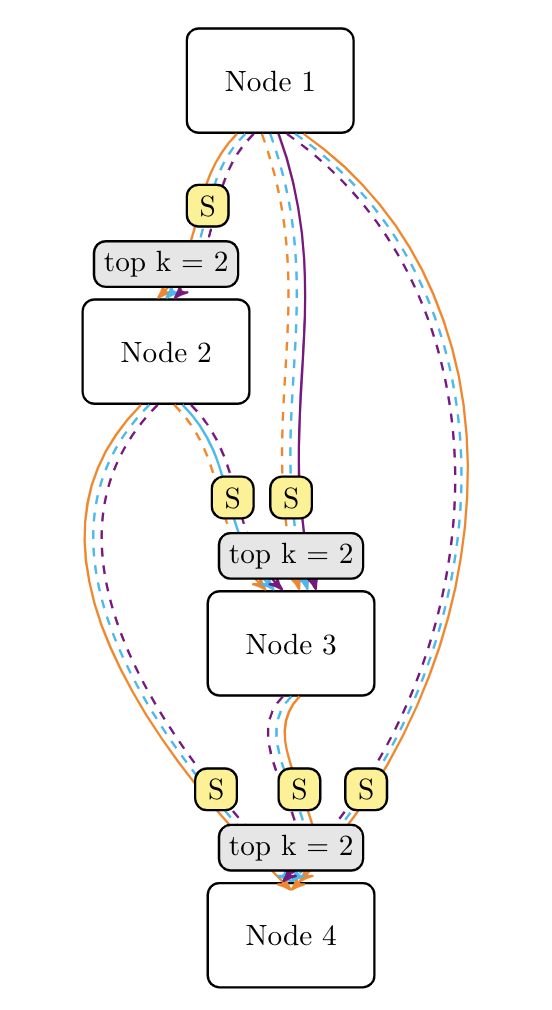}
\caption{Original DARTS \citep{liu2019darts}}
\label{fig-darts-cell}
\end{subfigure}
\begin{subfigure}{0.3\columnwidth}
\includegraphics[width=\columnwidth]{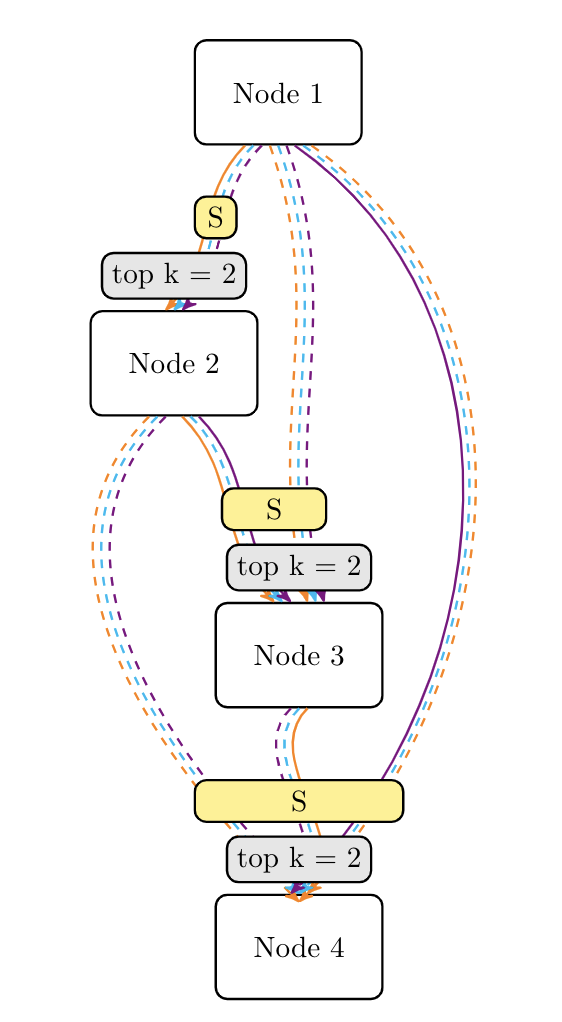}
\caption{I-DARTS \citep{idarts}}
\label{fig-idarts-cell}
\end{subfigure}
\begin{subfigure}{0.3\columnwidth}
\includegraphics[width=\columnwidth]{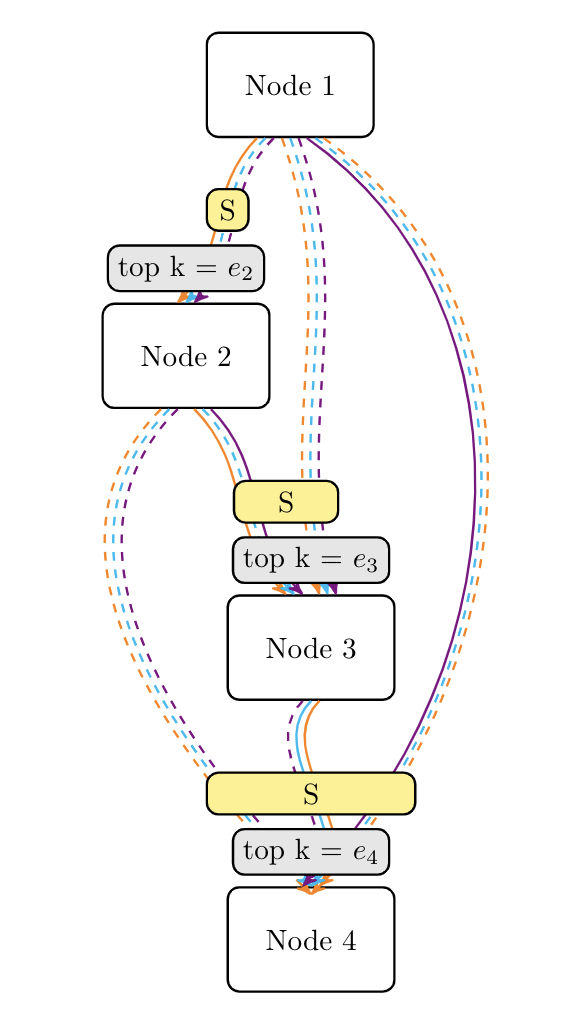}
\caption{XDARTS}
\label{fig-xdarts-cell}
\end{subfigure}
\caption{DARTS Cell Discretization Approaches}
\label{fig-darts-cells}
\end{center}
\end{figure*}

Next, we explored expanding the capacity of ICDARTS' cell search space by implementing new methods for discretization and incorporating them into the ICDARTS algorithm. As previously discussed, the search cells optimized in CDARTS and ICDARTS were initially introduced in DARTS. In these cells, a softmax operation is applied only to edges coming from the same node, and of these edges, only the one with the highest softmax values is eligible to be selected as an input to a cell (see Figure \ref{fig-darts-cell}). Allowing a maximum of one output of a previous node to be input to a current node eliminates any potential edge candidates from the same cell that may be more suitable than those from a different cell, thereby limiting the number of cell structures the search algorithm can discover. Based on this observation, \citep{idarts} presented the I-DARTS approach, in which all incoming edges to a node are compared equally by applying one softmax operation across all their weights. \(k\) edges are then selected from among these with equal weight, regardless of whether or not they come from different preceding nodes (see Figure \ref{fig-idarts-cell}). The advantages of this approach are that it increases the amount of cell architectures that can be discovered and allows all incoming edges to be compared fairly. After implementing a version of the I-DARTS cells and incorporating it into ICDARTS, we also designed and implemented a novel variant of this discretization approach, XDARTS, which further expands the search space of the cells. This XDARTS aims to compensate for any instability caused by selecting from a large set of edges at each depth. Following this discretization approach, the number of edges selected as inputs to a node equals the number of nodes preceding it. Thus, the number of edges selected at each node grows alongside its number of prospective input edges (see Figure \ref{fig-xdarts-cell}). 


\section{Experiments \& Results}
In this section, we will describe the experiments and results performed to evaluate the ICDARTS algorithm developed in this work.

\begin{figure*}[t]
\begin{center}

\includegraphics[width=0.45\columnwidth]{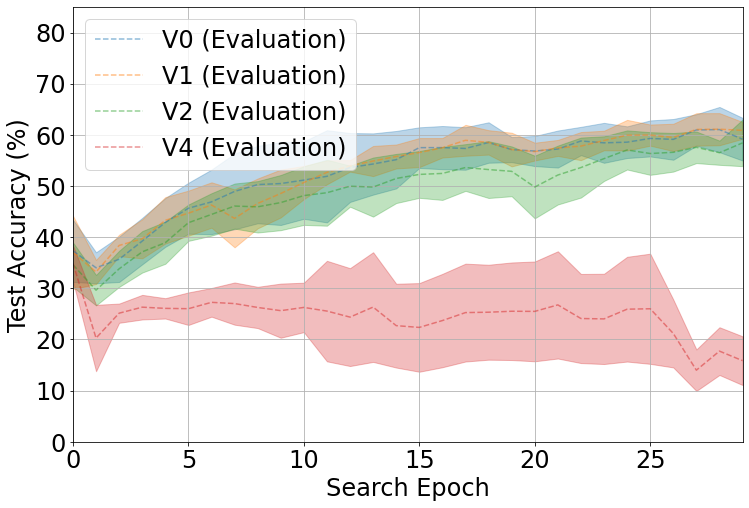}
\includegraphics[width=0.45\columnwidth]{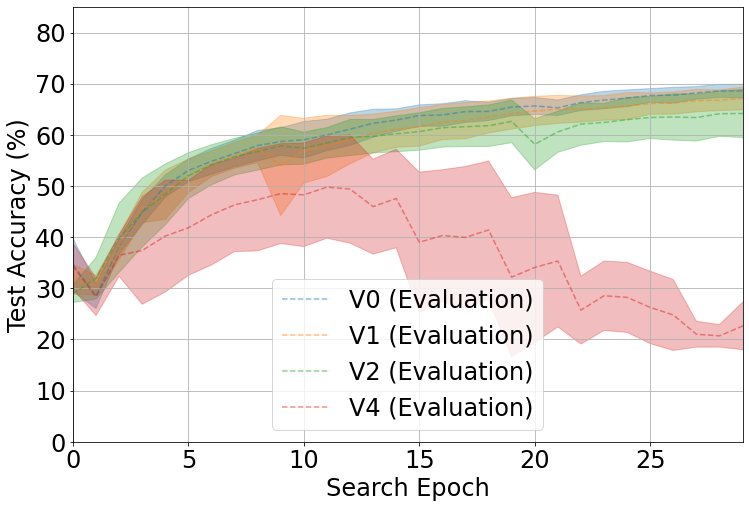}

\caption{CDARTS (left) and ICDARTS (right) Search Phase Evaluation Network CIFAR-10 Test Accuracy Curves Given Different \textit{Zero} Operation Configurations (see Table \ref{tab-zero-configs}).}
\label{fig-icdarts-results}
\end{center}
\end{figure*}
\begin{figure*}[ht]
\begin{center}
\centering
\resizebox{0.45\columnwidth}{!}{
\begin{tikzpicture}[->,>=stealth',auto,node distance=0.00cm,
  thick,main node/.style={circle,draw,font=\sffamily\Large\bfseries}]
\node[rectangle, fill=green!20, align=left, rounded corners, minimum width=1.0cm, minimum height=0.75cm, draw=black, yshift=-1.0cm] (c_k_1)  {\(c_{k-1}\)};
\node[rectangle, fill=green!20, align=left, rounded corners, minimum width=1.0cm, minimum height=0.75cm, draw=black, yshift=-4.0cm] (c_k_2) [below of=c_k_1]{\(c_{k-2}\)};
\node[rectangle, fill=cyan!20, align=left, rounded corners, minimum width=1.0cm, minimum height=0.75cm, draw=black, xshift=6.0cm, yshift=0cm] (n_0) [right of=c_k_1]{\(n_0\)};
\node[rectangle, fill=cyan!20, align=left, rounded corners, minimum width=1.0cm, minimum height=0.75cm, draw=black, xshift=6.0cm, yshift=-2cm] (n_1) [right of=c_k_1]{\(n_1\)};
\node[rectangle, fill=cyan!20, align=left, rounded corners, minimum width=1.0cm, minimum height=0.75cm, draw=black, xshift=6.0cm, yshift=-4cm] (n_2) [right of=c_k_1]{\(n_2\)};
\node[rectangle, fill=cyan!20, align=left, rounded corners, minimum width=1.0cm, minimum height=0.75cm, draw=black, xshift=10.0cm, yshift=0cm] (n_3) [right of=c_k_1]{\(n_3\)};
\node[rectangle, fill=yellow!20, align=left, rounded corners, minimum width=1.0cm, minimum height=0.75cm, draw=black, xshift=12.0cm, yshift=-2.0cm] (c_k) [right of=c_k_1]{\(c_k\)};

\draw [->] (c_k_1.east) to [bend left] node [above, align=center]  {sep\_conv\_3x3} (n_0.west);
\draw [->] (c_k_2.east) to [bend left] node [above, align=center, xshift=-0.1cm, yshift=0.5cm]  {sep\_conv\_3x3} (n_0.west);

\draw [->] (c_k_1.east) to [bend left] node [above, align=center, yshift=0.2cm]  {sep\_conv\_3x3} (n_1.west);
\draw [->] (c_k_2.east) to [bend right] node [above, align=center, xshift=-1.2cm, yshift=-0.2cm]  {sep\_conv\_3x3} (n_1.west);

\draw [->] (c_k_1.east) to [bend right] node [above, align=center, xshift=0.7cm, yshift=0.2cm]  {sep\_conv\_3x3} (n_2.west);
\draw [->] (c_k_2.east) to [bend right] node [above, align=center, yshift=0.1cm]  {sep\_conv\_3x3} (n_2.west);

\draw [->] (c_k_1.east) to [out=55,in=150] node [above, align=center]  {sep\_conv\_3x3} (n_3.north);
\draw [->] (n_0.east) to node [above, align=center]  {sep\_conv\_3x3} (n_3.west);

\draw [->] (n_3.east) to  (c_k.north);
\draw [->] (n_0.east) to  (c_k.west);
\draw [->] (n_1.east) to (c_k.west);
\draw [->] (n_2.east) to (c_k.west);

\end{tikzpicture}}
\resizebox{0.45\columnwidth}{!}{
\begin{tikzpicture}[->,>=stealth',auto,node distance=0.00cm,
  thick,main node/.style={circle,draw,font=\sffamily\Large\bfseries}]

\node[rectangle, fill=green!20, align=left, rounded corners, minimum width=1.0cm, minimum height=0.75cm, draw=black, yshift=-1.0cm] (c_k_1)  {\(c_{k-1}\)};
\node[rectangle, fill=green!20, align=left, rounded corners, minimum width=1.0cm, minimum height=0.75cm, draw=black, yshift=-4.0cm] (c_k_2) [below of=c_k_1]{\(c_{k-2}\)};
\node[rectangle, fill=cyan!20, align=left, rounded corners, minimum width=1.0cm, minimum height=0.75cm, draw=black, xshift=3.0cm, yshift=-3cm] (n_0) [right of=c_k_1]{\(n_0\)};
\node[rectangle, fill=cyan!20, align=left, rounded corners, minimum width=1.0cm, minimum height=0.75cm, draw=black, xshift=6.5cm, yshift=-4cm] (n_1) [right of=c_k_1]{\(n_1\)};
\node[rectangle, fill=cyan!20, align=left, rounded corners, minimum width=1.0cm, minimum height=0.75cm, draw=black, xshift=6.0cm, yshift=0cm] (n_2) [right of=c_k_1]{\(n_2\)};
\node[rectangle, fill=cyan!20, align=left, rounded corners, minimum width=1.0cm, minimum height=0.75cm, draw=black, xshift=9.0cm, yshift=-2cm] (n_3) [right of=c_k_1]{\(n_3\)};
\node[rectangle, fill=yellow!20, align=left, rounded corners, minimum width=1.0cm, minimum height=0.75cm, draw=black, xshift=12.0cm, yshift=-2.0cm] (c_k) [right of=c_k_1]{\(c_k\)};

\draw [->] (c_k_1.east) to node [above, align=center, xshift=1.3cm, yshift=-0.4cm]  {avg\_pool\_3x3} (n_0.west);
\draw [->] (c_k_2.east) to node [above, align=center, xshift=-0.5cm, yshift=0.1cm]  {identity} (n_0.west);

\draw [->] (c_k_1.east) to [bend left] node [above, align=center, xshift=-0.5cm, yshift=0.8cm]  {sep\_conv\_3x3} (n_1.west);
\draw [->] (n_0.east) to [bend left] node [above, align=center, xshift=-0.5cm, yshift=-0.8cm]  {sep\_conv\_5x5} (n_1.west);

\draw [->] (n_0.east) to  node [above, align=center, xshift=1.4cm, yshift=0.0cm]  {dil\_conv\_5x5} (n_2.south);
\draw [->] (c_k_1.east) to [bend left] node [above, align=center, yshift=0.1cm]  {sep\_conv\_3x3} (n_2.west);

\draw [->] (n_2.east) to node [above, align=center, xshift=1.2cm, yshift=-0.2cm]  {dil\_conv\_5x5} (n_3.west);
\draw [->] (n_1.east) to node [above, align=center, xshift=1.2cm, yshift=-0.2cm]  {dil\_conv\_5x5} (n_3.west);

\draw [->] (n_3.east) to  (c_k.west);
\draw [->] (n_0.south) to [out=-80,in=-100] (c_k.west);
\draw [->] (n_1.east) to [bend right] (c_k.west);
\draw [->] (n_2.east) to [bend left] (c_k.west);

\end{tikzpicture}}
\caption{CDARTS V0 normal (left) and reduction (right) cells from the network with the best overall test accuracy.}
\label{fig-cdarts-example}
\end{center}
\end{figure*}
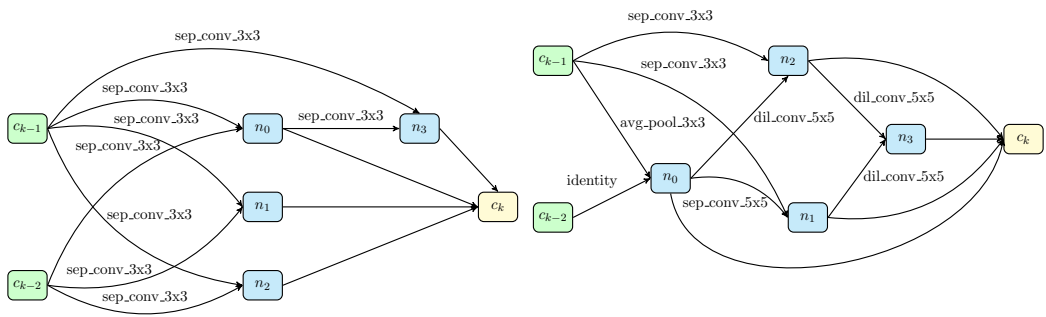
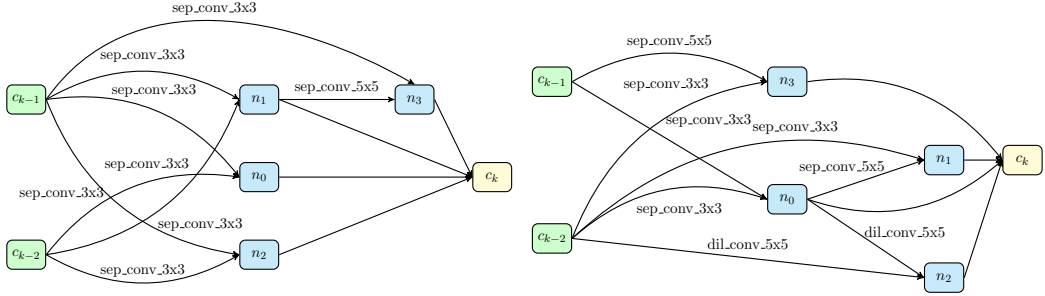
\begin{figure*}[ht]
\begin{center}
\centering
\resizebox{0.45\columnwidth}{!}{
\begin{tikzpicture}[->,>=stealth',auto,node distance=0.00cm,
  thick,main node/.style={circle,draw,font=\sffamily\Large\bfseries}]

\node[rectangle, fill=green!20, align=left, rounded corners, minimum width=1.0cm, minimum height=0.75cm, draw=black, yshift=-1.0cm] (c_k_1)  {\(c_{k-1}\)};
\node[rectangle, fill=green!20, align=left, rounded corners, minimum width=1.0cm, minimum height=0.75cm, draw=black, yshift=-4.0cm] (c_k_2) [below of=c_k_1]{\(c_{k-2}\)};
\node[rectangle, fill=cyan!20, align=left, rounded corners, minimum width=1.0cm, minimum height=0.75cm, draw=black, xshift=6.0cm, yshift=-2cm] (n_0) [right of=c_k_1]{\(n_0\)};
\node[rectangle, fill=cyan!20, align=left, rounded corners, minimum width=1.0cm, minimum height=0.75cm, draw=black, xshift=6.0cm, yshift=0cm] (n_1) [right of=c_k_1]{\(n_1\)};
\node[rectangle, fill=cyan!20, align=left, rounded corners, minimum width=1.0cm, minimum height=0.75cm, draw=black, xshift=6.0cm, yshift=-4cm] (n_2) [right of=c_k_1]{\(n_2\)};
\node[rectangle, fill=cyan!20, align=left, rounded corners, minimum width=1.0cm, minimum height=0.75cm, draw=black, xshift=10.0cm, yshift=0cm] (n_3) [right of=c_k_1]{\(n_3\)};
\node[rectangle, fill=yellow!20, align=left, rounded corners, minimum width=1.0cm, minimum height=0.75cm, draw=black, xshift=12.0cm, yshift=-2.0cm] (c_k) [right of=c_k_1]{\(c_k\)};

\draw [->] (c_k_1.east) to [bend left] node [above, align=center, xshift=0.0cm, yshift=0.2cm]  {sep\_conv\_3x3} (n_0.west);
\draw [->] (c_k_2.east) to [bend left] node [above, align=center, xshift=0.4cm, yshift=0.3cm]  {sep\_conv\_3x3} (n_0.west);

\draw [->] (c_k_1.east) to [bend left] node [above, align=center, yshift=0.1cm]  {sep\_conv\_3x3} (n_1.west);
\draw [->] (c_k_2.east) to [bend right] node [above, align=center, xshift=0.9cm, yshift=-0.8cm]  {sep\_conv\_3x3} (n_1.west);

\draw [->] (c_k_1.east) to [bend right] node [above, align=center, xshift=-1.5cm, yshift=0.0cm]  {sep\_conv\_3x3} (n_2.west);
\draw [->] (c_k_2.east) to [bend right] node [above, align=center, yshift=0.0cm]  {sep\_conv\_3x3} (n_2.west);

\draw [->] (c_k_1.east) to [out=55,in=150] node [above, align=center]  {sep\_conv\_3x3} (n_3.north);
\draw [->] (n_1.east) to node [above, align=center]  {sep\_conv\_5x5} (n_3.west);

\draw [->] (n_3.east) to  (c_k.west);
\draw [->] (n_0.east) to  (c_k.west);
\draw [->] (n_1.east) to (c_k.west);
\draw [->] (n_2.east) to (c_k.west);

\end{tikzpicture} } 
\resizebox{0.45\columnwidth}{!}{
\begin{tikzpicture}[->,>=stealth',auto,node distance=0.00cm,
  thick,main node/.style={circle,draw,font=\sffamily\Large\bfseries}]

\node[rectangle, fill=green!20, align=left, rounded corners, minimum width=1.0cm, minimum height=0.75cm, draw=black, yshift=-1.0cm] (c_k_1)  {\(c_{k-1}\)};
\node[rectangle, fill=green!20, align=left, rounded corners, minimum width=1.0cm, minimum height=0.75cm, draw=black, yshift=-4.0cm] (c_k_2) [below of=c_k_1]{\(c_{k-2}\)};
\node[rectangle, fill=cyan!20, align=left, rounded corners, minimum width=1.0cm, minimum height=0.75cm, draw=black, xshift=6.0cm, yshift=-3.0cm] (n_0) [right of=c_k_1]{\(n_0\)};
\node[rectangle, fill=cyan!20, align=left, rounded corners, minimum width=1.0cm, minimum height=0.75cm, draw=black, xshift=10.0cm, yshift=-2cm] (n_1) [right of=c_k_1]{\(n_1\)};
\node[rectangle, fill=cyan!20, align=left, rounded corners, minimum width=1.0cm, minimum height=0.75cm, draw=black, xshift=10.0cm, yshift=-5cm] (n_2) [right of=c_k_1]{\(n_2\)};
\node[rectangle, fill=cyan!20, align=left, rounded corners, minimum width=1.0cm, minimum height=0.75cm, draw=black, xshift=6.0cm, yshift=0cm] (n_3) [right of=c_k_1]{\(n_3\)};
\node[rectangle, fill=yellow!20, align=left, rounded corners, minimum width=1.0cm, minimum height=0.75cm, draw=black, xshift=12.0cm, yshift=-2.0cm] (c_k) [right of=c_k_1]{\(c_k\)};

\draw [->] (c_k_1.east) to node [above, align=center, xshift=1.0cm, yshift=0.2cm]  {sep\_conv\_3x3} (n_0.west);
\draw [->] (c_k_2.east) to [bend left] node [below, align=center, xshift=0.4cm, yshift=-0.2cm]  {sep\_conv\_3x3} (n_0.west);

\draw [->] (c_k_2.east) to [bend left] node [above, align=center, xshift=1.5cm, yshift=0.2cm]  {sep\_conv\_3x3} (n_1.west);
\draw [->] (n_0.east) to node [above, align=center, xshift=-0.6cm, yshift=-0cm]  {sep\_conv\_5x5} (n_1.west);

\draw [->] (n_0.east) to node [above, align=center, xshift=1.0cm, yshift=-0.1cm]  {dil\_conv\_5x5} (n_2.west);
\draw [->] (c_k_2.east) to node [above, align=center, yshift=0.1cm]  {dil\_conv\_5x5} (n_2.west);

\draw [->] (c_k_1.east) to [bend left] node [above, align=center, xshift=0.0cm, yshift=0.0cm]  {sep\_conv\_5x5} (n_3.west);
\draw [->] (c_k_2.east) to [bend left] node [above, align=center, xshift=0.5cm, yshift=0.9cm]  {sep\_conv\_3x3} (n_3.west);

\draw [->] (n_3.east) to [bend left] (c_k.west);
\draw [->] (n_0.east) to [bend right] (c_k.west);
\draw [->] (n_1.east) to (c_k.west);
\draw [->] (n_2.east) to (c_k.west);

\end{tikzpicture}}
\caption{ICDARTS V1 normal (left) and reduction (right) cells from the network with the best overall test accuracy.}
\label{fig-icdarts-example}
\end{center}
\end{figure*}

\begin{table}[h]
\caption{CIFAR-10 Retraining Test Set Accuracies of Networks Produced by CDARTS, ICDARTS, and Random Initialization using Different \textit{Zero} Return Configurations (see Table \ref{tab-zero-configs}).}
\begin{center}
\begin{tabular}{|c|c|c|c|c|}
\hline
\textbf{Zero Return}&\multicolumn{3}{|c|}{\textbf{Algorithm}} \\
\cline{2-4} 
\textbf{Configuration} & \textbf{\textit{Random}}& \textbf{\textit{CDARTS}}& \textbf{\textit{ICDARTS}} \\ 
\hline
V0  & 96.78 (0.39)  & 97.17 (0.21) &  96.99 (\textbf{0.16})\\
\hline
V1  & 96.57 (0.32) & 97.01 (0.24) & 97.10 (\textbf{0.19})  \\
\hline
V2  & 96.52 (0.37) & 96.97 (0.27) & 96.92 (\textbf{0.15}) \\
\hline
V3  & 96.58 (0.31) & 96.96 (0.28) & 96.93 (\textbf{0.14}) \\
\hline
V4  & 96.58 (0.31) & 23.68 (29.71)  & 90.00 (\textbf{12.72}) \\
\hline
\end{tabular}
\label{tab-cifar10-results}
\end{center}
\end{table}

\begin{table}[h]
\caption{CIFAR-100 Retraining Test Set Accuracies of Networks Produced by CDARTS, ICDARTS, and Random Initialization using Different \textit{Zero} Return Configurations (see Table \ref{tab-zero-configs}).}
\begin{center}
\begin{tabular}{|c|c|c|c|c|}
\hline
\textbf{Zero Return}&\multicolumn{3}{|c|}{\textbf{Algorithm}} \\
\cline{2-4} 
\textbf{Configuration} & \textbf{\textit{Random}}& \textbf{\textit{CDARTS}}& \textbf{\textit{ICDARTS}} \\ 
\hline 
V0  & 81.11 (1.39) & 83.66 (\textbf{0.29}) & 83.10 (0.65) \\
\hline
V1 & 80.54 (1.05) & 83.13 (0.71) & 83.33 (\textbf{0.19}) \\
\hline
V2 & 80.63 (1.26) & 83.19 (0.68) & 83.19 (\textbf{0.21})\\
\hline
V3 & 80.84 (1.13) & 83.05 (0.72) & 83.15 (\textbf{0.36}) \\
\hline
V4 & 80.84 (1.13) & 10.98 (\textbf{25.91}) & 63.13 (28.35) \\
\hline
\end{tabular}
\label{tab-cifar100-results}
\end{center}
\end{table}

\subsection{Datasets}
For each change to the ICDARTS algorithm and search space, the architecture search phase is conducted on the CIFAR-10 dataset and the resulting network architecture is retrained on CIFAR-10. The networks produced before and after the initial algorithmic changes that resulted in the ICDARTS algorithm, as well as the experiments on the \textit{zero} operation, were additionally retrained and evaluated on the CIFAR-100 dataset. The CIFAR-10 and CIFAR-100 image classification benchmarks consist of $10$ and $100$ classes, respectively, and both are comprised of $50K$ training and $10K$ testing images of resolution $32 \times 32$.

\subsection{Search \& Evaluation Settings}
At the beginning of the search phase, the original training set is divided into two datasets of equal size, denoted by \textit{train} and \textit{val}, as in \citep{liu2019darts} and \citep{yu2020cyclic}. As previously discussed, in the ICDARTS algorithm, the \textit{train} partition is used for updating the weights of both the search and evaluation networks, while \textit{val} is reserved for updating the $\alpha$ weights.
The search phase runs for 30 epochs, not including 2 epochs for pre-training the search network and 1 for warming up the intermediate evaluation network each time it is generated. The search and evaluation weights, \(w_S\) and \(w_E\) are updated using separate SGD optimizers with learning rates of \(0.08\), decay rates of \(3 \times 10^{-4}\), and momentum settings of 0.9. The \(\alpha\) weights are updated using an Adam optimizer \citep{kingma2017adam} with a learning rate of \(3 \times 10^{-4}\), decay rate of \(0\), and momentum \(\beta\)s of \({0.5, 0.999}\). 

The evaluation stage involves retraining the discovered architectures for 600 epochs on the full training dataset and evaluating the retrained network on the test dataset. Our retraining procedure closely resembles that of the original CDARTS paper. The batch size is set to 128, and an SGD optimizer is employed with a learning rate of \(0.025\), momentum of \(0.9\), and weight decay of \(5 \times 10^{-4}\). As in the search phase, this optimizer is paired with a cosine annealing learning rate scheduler. Following the approach of \citep{pham2018efficient,zoph2018learning,liu2018progressive}, the training dataset is augmented with a cutout regulation length of 16 \citep{devries2017improved}, the drop path rate of the full evaluation network is set to \(0.3\), and the auxiliary towers to \(0.4\). 
The search and evaluation phases of each experiment discussed in this publication were each run 8 times unless indicated otherwise in their result tables.

\subsection{ICDARTS}

Figure \ref{fig-icdarts-results} depicts the test set accuracies of the evaluation networks throughout the search phases of CDARTS and ICDARTS. Each curve has been plotted with a \(95\%\) confidence interval across the runs of each configuration. Search network pre-training epochs are not included in these graphs. The lower standard deviation of the revised algorithm is evidence that our approach gives results with improved consistency and stability across different initializations. The accuracy disparity between the search and retraining phases is expected due to the differences in data augmentation and the number of epochs trained.
 
The evaluation network curve for the V4 search space configuration, which includes the \textit{zero} operation in the search phase evaluation network but not the search network, shows particularly poor stability and difficulty learning. This outcome is likely the product of the difference in behavior between the search and evaluation networks during the search process, which results in a feedback loop that produces an increasing number of \(zero\) operations in the discretized evaluation network.

Tables \ref{tab-cifar10-results} and \ref{tab-cifar100-results} list the average and standard deviation test set accuracies of the evaluation networks on CIFAR-10 and CIFAR-100. Note that the values we report were obtained at the end of the full retraining cycle rather than the best test performance obtained at any point during retraining( as was reported in \citep{liu2019darts} and later \citep{yu2020cyclic}). Hence, our results measure a typical run's performance rather than report our best outlier's performance. The accuracy results of the architecture produced randomly and by the CDARTS and ICDARTS algorithms show that the ICDARTS networks achieved similar mean accuracies to that of the CDARTS network but with much smaller variation in performance, demonstrating stability improvements and results that are potentially more reproducible.

    \begin{table}[h]
    \caption{ICDARTS Template Ablation Study Retraining CIFAR-10 Test Set Accuracies and Inference Latencies}
    \begin{center}
    \begin{tabular}{|c|c|c|}
    \hline
    \textbf{Ablation}&\multicolumn{2}{|c|}{\textbf{Results}} \\
    \cline{2-3} & \textbf{Retraining Accuracy} & \textbf{Inference Latency (batch/s)}\\ 
    \hline
    Original & 97.13 (0.14) & 0.10 (0.01)\\
    \hline
    Pooling & \textbf{97.19 (0.23)} & 0.09 (0.02)\\
    \hline
    Identity & 96.96 (0.35) & 0.11 (0.02) \\
    \hline
    Dilated Convolutions  & 97.02 (0.26) & 0.10 (0.01) \\
    \hline
    Separable Convolutions  & 96.15 (0.27) & \textbf{0.07 (0.00)}\\
    \hline
    Auxilary Heads \(^*\) & 96.86 (0.14) & 0.11 (0.00)\\
    \hline
    Stem Identity \(^*\)  & 80.01 (2.18) & 0.09 (0.02)\\
    \hline
    Stem Conv BN \(^*\) & 96.81 (0.38) & 0.11 (0.03)\\
    \hline
    Stem Concat \(^*\) & 92.40 (0.33) & 0.16 (0.00) \\
    \hline
    Reduce AvgPool & 96.94 (0.19) & 0.12 (0.07)\\
    \hline
    Reduce MaxPool & 96.87 (0.06) &  0.10 (0.03)\\
    \hline
    Reduce Convolution & 96.77 (0.12) & 0.09 (0.01) \\
    \hline
    \multicolumn{3}{l}{Note: * indicates results that are averages of 4 runs.}\\
    \end{tabular}
    \label{tab-template-ablation-results}
    \end{center}
    \end{table}

\begin{figure*}[t]
\begin{center}
\includegraphics[width=0.45\columnwidth]{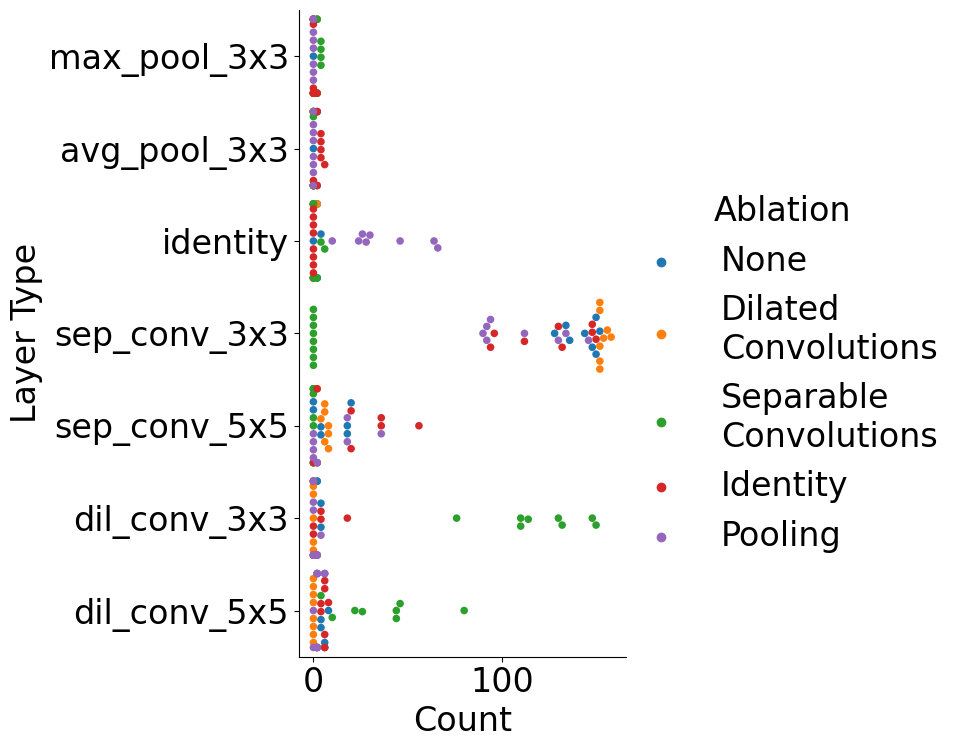}\\
\vspace{0.1cm}
\includegraphics[width=0.45\columnwidth]{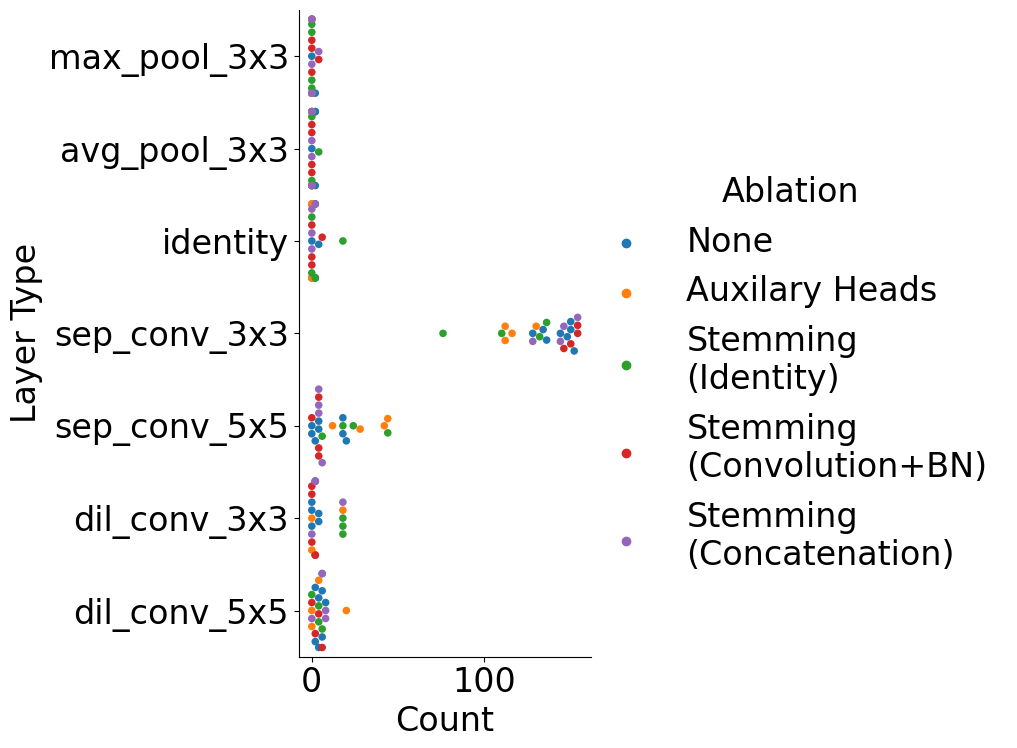}
\vspace{0.1cm}
\includegraphics[width=0.45\columnwidth]{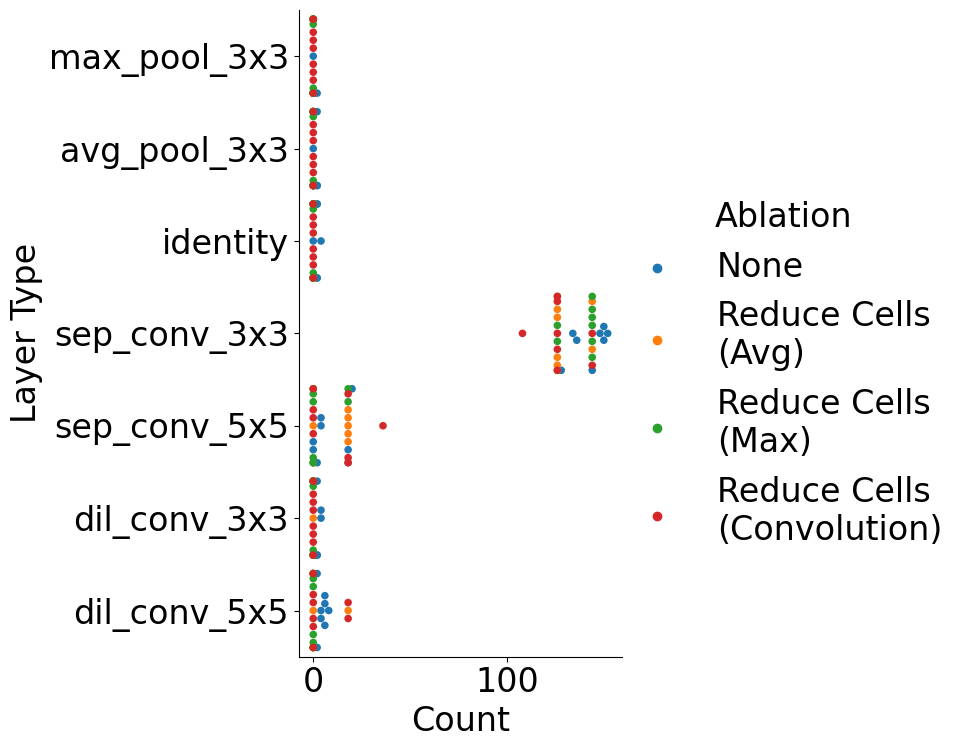}
\caption{ICDARTS Template Ablation Per Network Layer Type Frequencies}
\label{fig-template-ablation-catplot}
\end{center}
\end{figure*}

\begin{figure*}[t]
\begin{center}
\includegraphics[width=0.4\columnwidth]{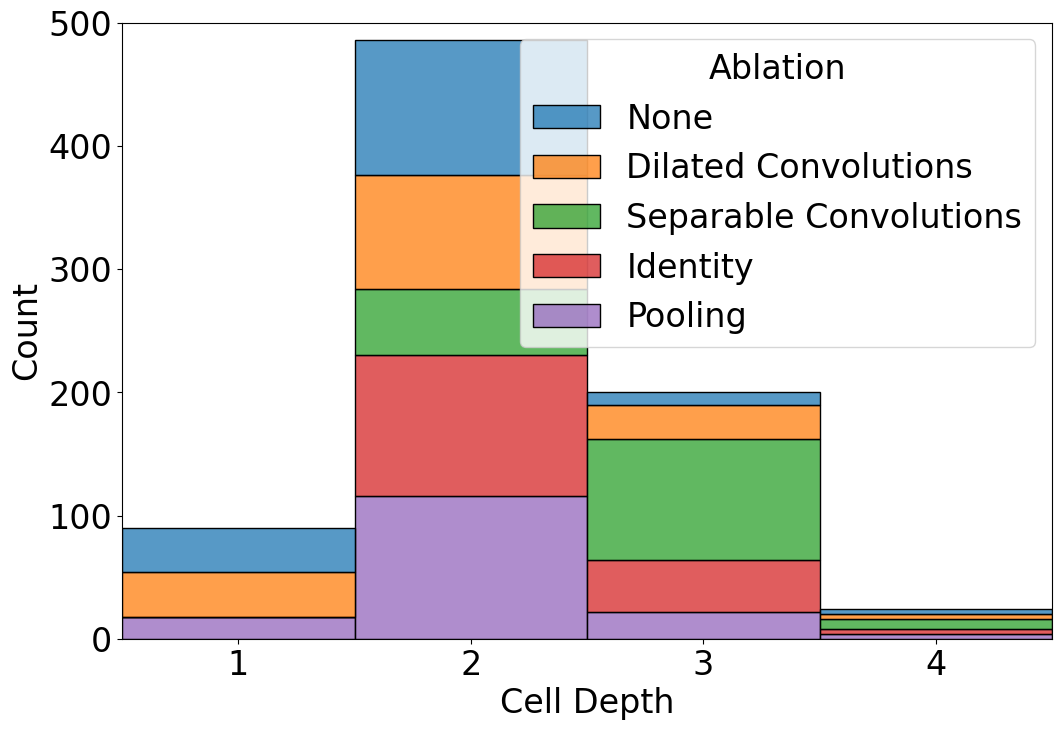}\\
\vspace{0.1cm}
\includegraphics[width=0.4\columnwidth]{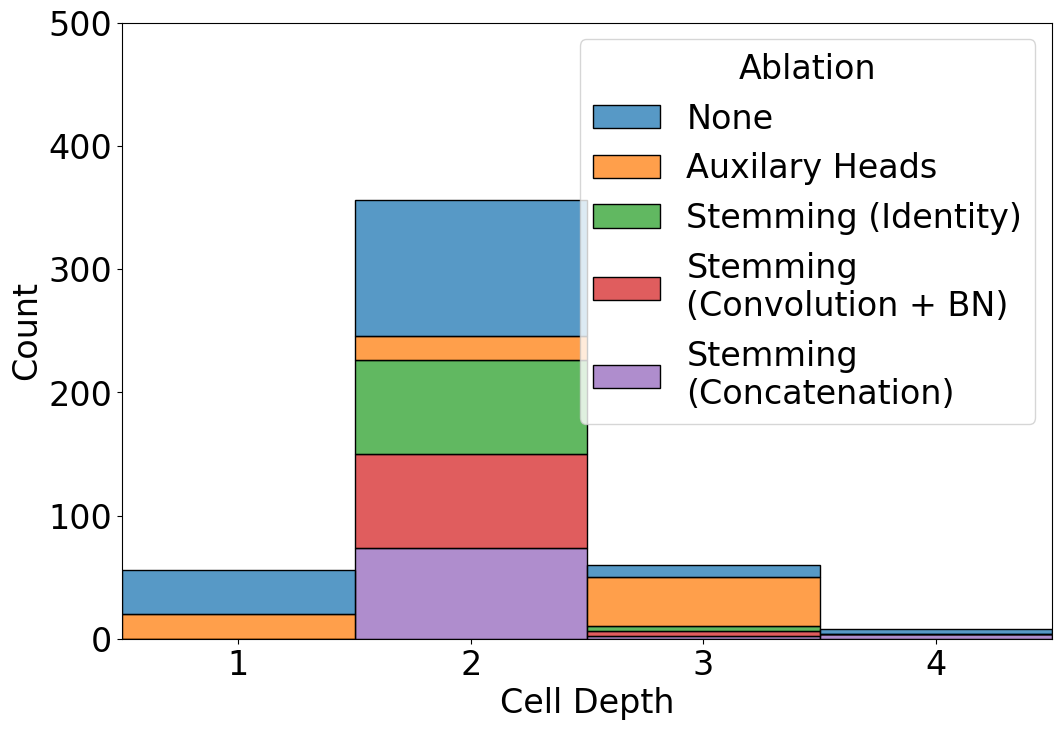}
\vspace{0.1cm}
\includegraphics[width=0.4\columnwidth]{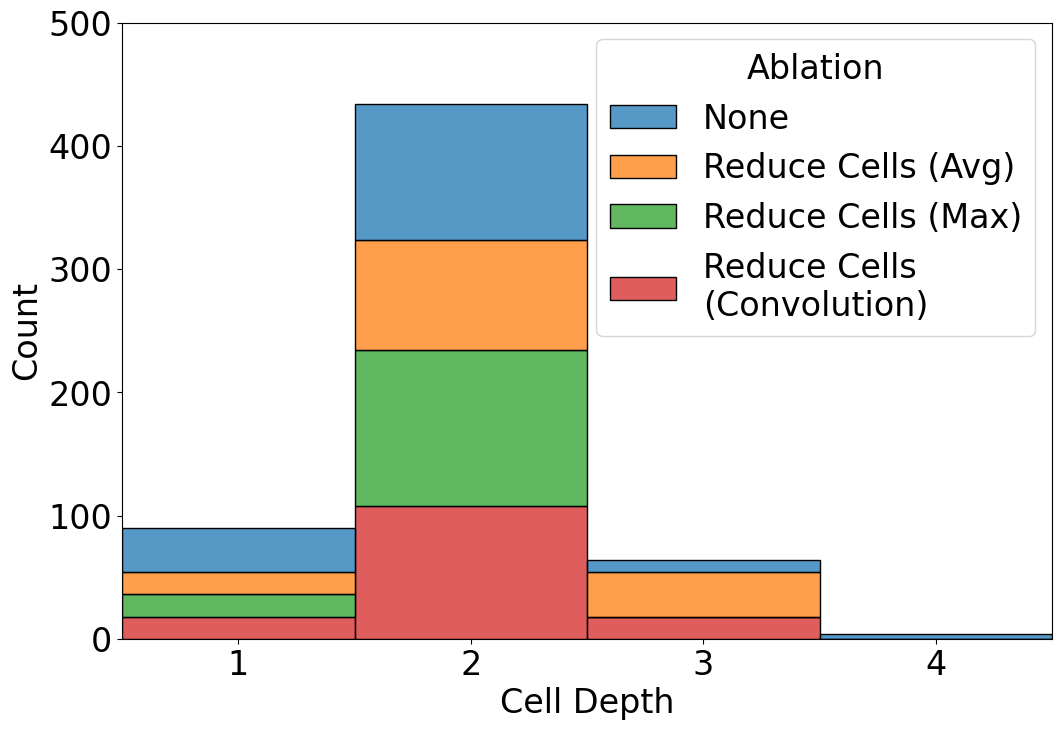}
\caption{ICDARTS Template Ablation Per Network Cell Depth Frequencies}
\label{fig-template-ablation-cell-depths}
\end{center}
\end{figure*}

\subsection{Ablation Studies}

Table \ref{tab_temp_ablation} and Figures \ref{fig-template-ablation-catplot} and \ref{fig-template-ablation-cell-depths} show the results of the ablation study of the ICDARTS template. Note the inference latencies were obtained by calculating the average per batch inference times on the test dataset prior to re-training. Additionally, the operation type and cell depth frequencies shown in Figures \ref{fig-template-ablation-catplot} and \ref{fig-template-ablation-cell-depths} are represented as the totals in each evaluation network in order to account for the difference in the number of normal and reduce cells in each network. For the cell operation choice ablations, only the ablation of the pooling operations improved the average retraining accuracy, likely because this resulted in less competition with better-performing operations and the selection of more identity operations (see Figure \ref{fig-template-ablation-catplot}). The retraining accuracies for the rest of the ablations fell below that of the original, with the ablation of the separable convolutions yielding the lowest accuracies for this category of ablation. On the other hand, each cell operation choice ablation improved average latencies, except for the dilated convolution ablation. This trend likely occurred because removing this option caused the algorithm to favor the computationally expensive separable convolutions, as shown in Figure \ref{fig-template-ablation-catplot}. The ablation of the separable convolution operations improved the retraining latencies by the most significant margin, although it also produced the worst average retraining accuracy.  

The ablation of the auxiliary heads also resulted in lower retraining accuracies. However, this ablation has higher latencies, a finding that might be explained by this ablation resulting in ICDARTS favoring a deeper cell structure (see Figure \ref{fig-template-ablation-cell-depths}).

The stemming layer ablations also offered no improvement in retraining accuracies and worse inference latencies. The only exception was when the layer was replaced with an identity operation so that the number of input channels was equal to that of the input images, which resulted in a network with fewer parameters and, hence, a lower latency. 

Finally, the accuracies from the reduce cell ablations were also worse than that of ICDARTS. On the other hand, the average inference latencies varied, with the case in which reduce cells were replaced with convolutions with stride two providing the only latency improvement.

The results of the algorithmic ablation study listed in Table \ref{tab-alg_ablation} generally demonstrate stability improvements with the addition of each modification to the original algorithm on both routes. The inference latencies tend to slow with each change. The exception to this pattern was route A, in which the switch to updating both network weights on the training set was not made until the final step. However, as shown in the tables, the slower latencies were eventually rectified with the final modification to the algorithm on both routes.

\begin{table}[h]
\caption{ICDARTS Algorithmic Ablation CIFAR-10 Retraining Test Set Accuracies and Inference Latencies. The details of each ablation are listed in Table \ref{tab-alg-ablation-routes}.}
\begin{center}
\resizebox{\columnwidth}{!}{%
\begin{tabular}{|c|c|c|}
\hline
\textbf{Ablation}&  \textbf{Retraining Accuracy} & \textbf{Inference Latency (batch/s)}\\ 
\hline
CDARTS (No \textit{Zero} Operation) & 96.94 (0.30)	& 0.09 (0.01) \\
\hline
A1 & 96.35 (0.48) & \textbf{0.08 (0.00)}\\
\hline
B1 & 96.99 (0.22) & 0.09 (0.01)\\
\hline
A2 & 96.69 (0.22) &  \textbf{0.08 (0.00)} \\
\hline
B2 & 97.05 (0.18) &  0.11 (0.00) \\
\hline
ICDARTS & \textbf{97.13 (0.14)} & 0.10 (0.01)\\
\hline
\end{tabular}
}
\label{tab-alg_ablation}
\end{center}
\end{table}

\begin{figure*}[t]
  
\begin{center}
\centerline{
\includegraphics[width=0.45\columnwidth]{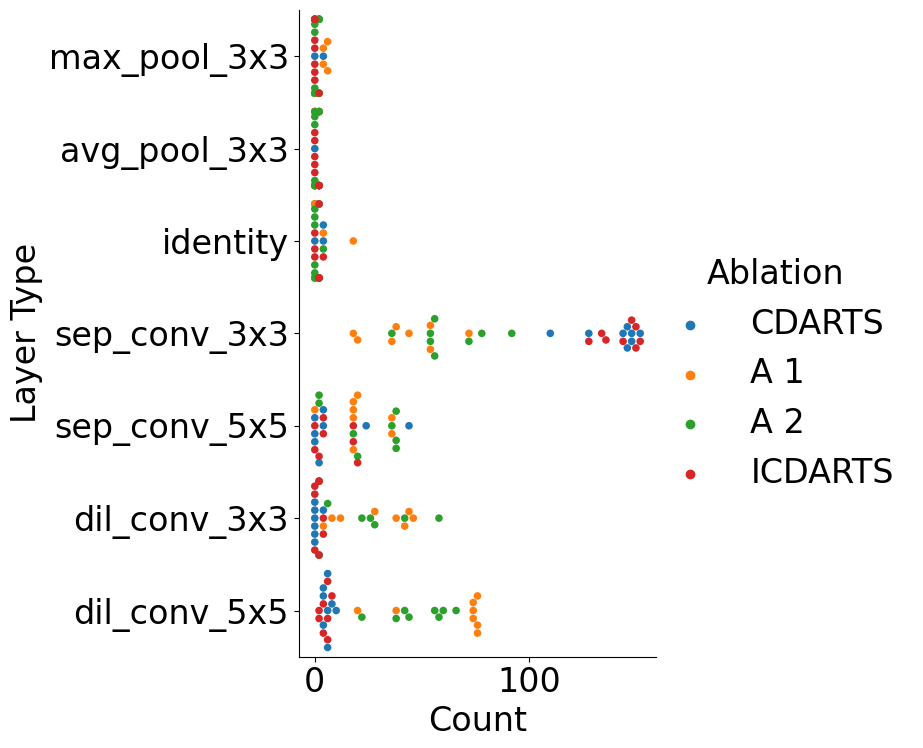}
\includegraphics[width=0.45\columnwidth]{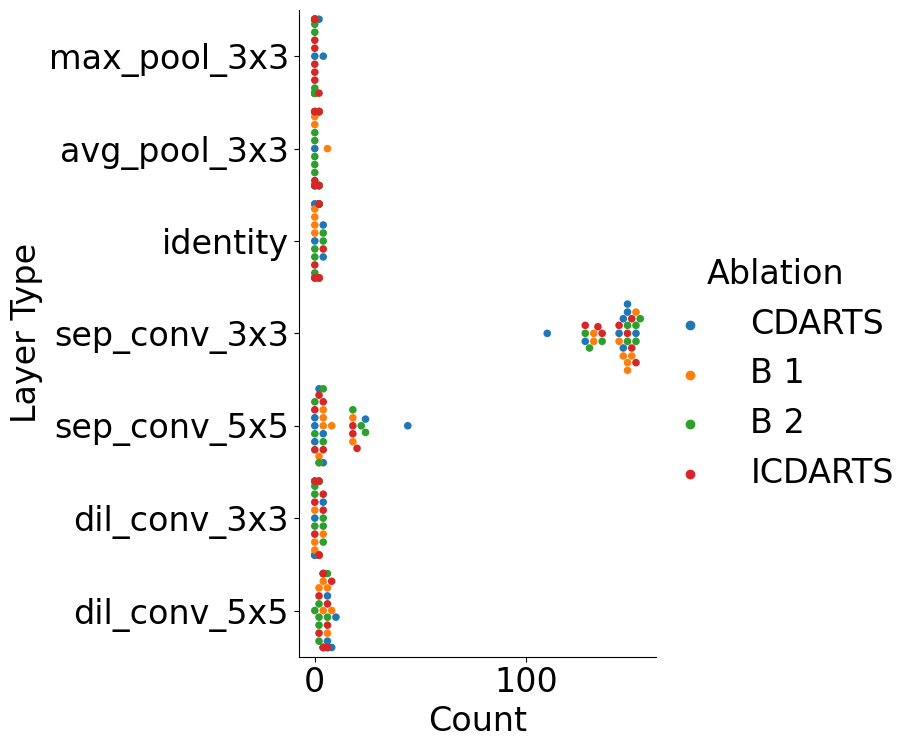}
}
\caption{ICDARTS Per Network Layer Type Frequencies for Algorithmic Ablation Study Routes A (left) and B (right). The details of each route are listed in Table \ref{tab-alg-ablation-routes}.}
\label{fig-alg-ablation-catplot}
\end{center}
\end{figure*}

\begin{figure*}[t]
  
\begin{center}
\centerline{
\includegraphics[width=0.45\columnwidth]{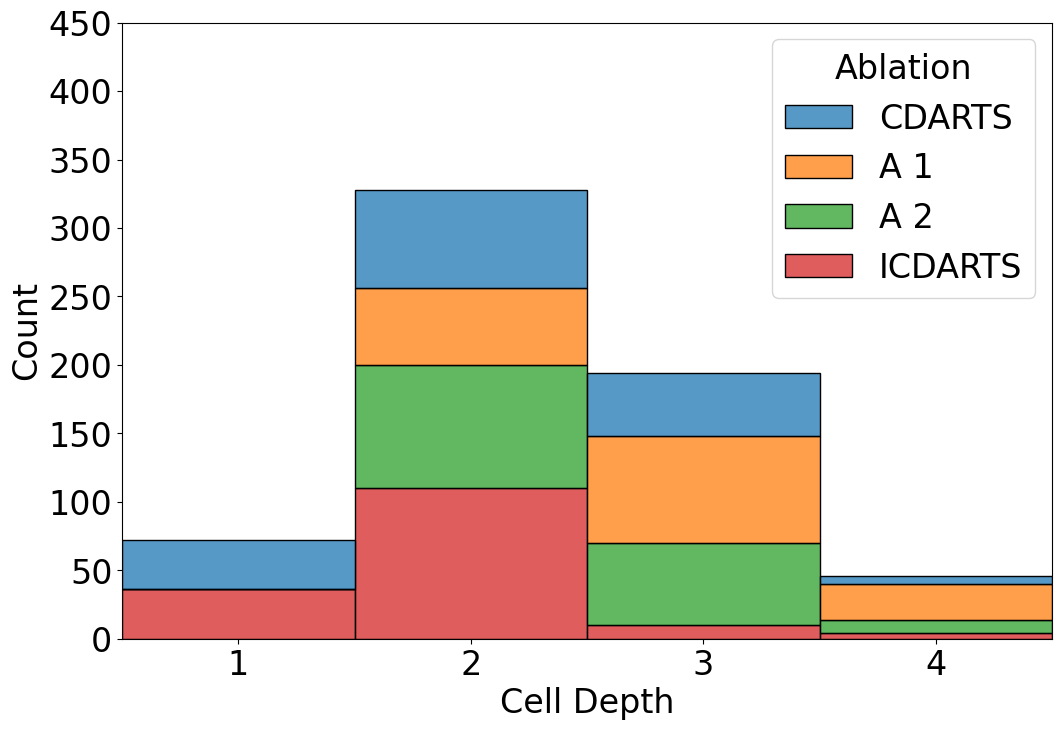}

\includegraphics[width=0.45\columnwidth]{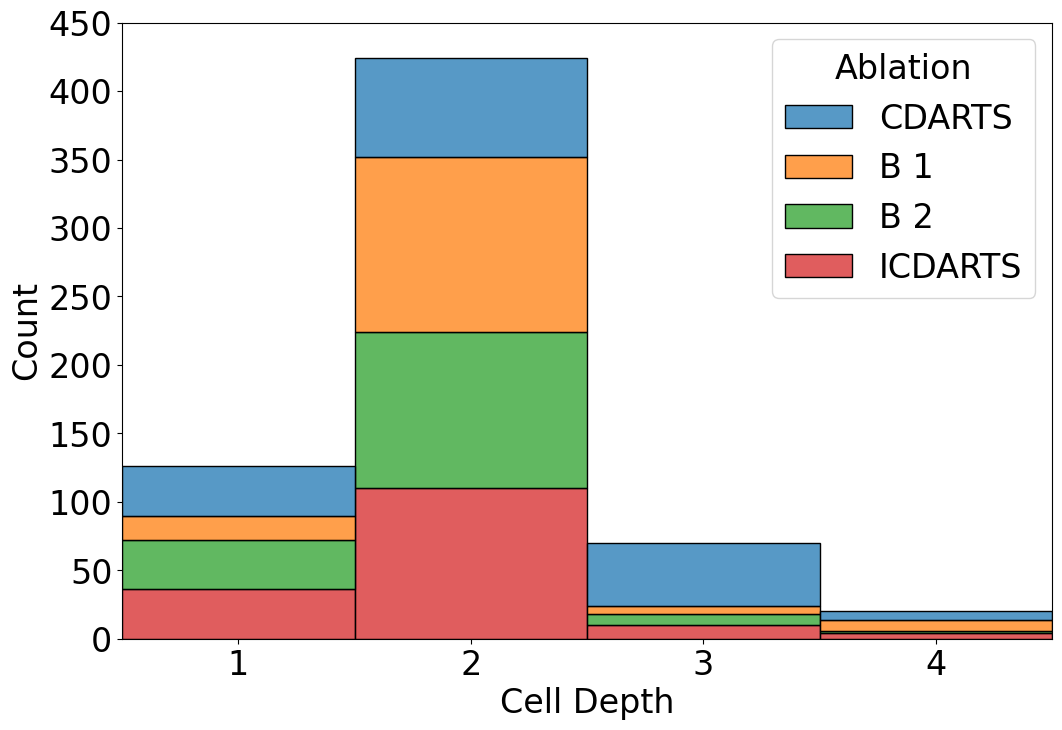}}
\caption{ICDARTS Per Network Cell Depth Frequencies for Algorithmic Ablation Study Routes A (left) and B (right). The detail of each route are found in Table \ref{tab-alg-ablation-routes}.}
\label{fig-alg-ablation-cell-depths}
\end{center}
\end{figure*}

\subsection{Alternate Search Spaces}

\begin{figure*}[t]
  
\begin{center}
\includegraphics[width=0.45\columnwidth]{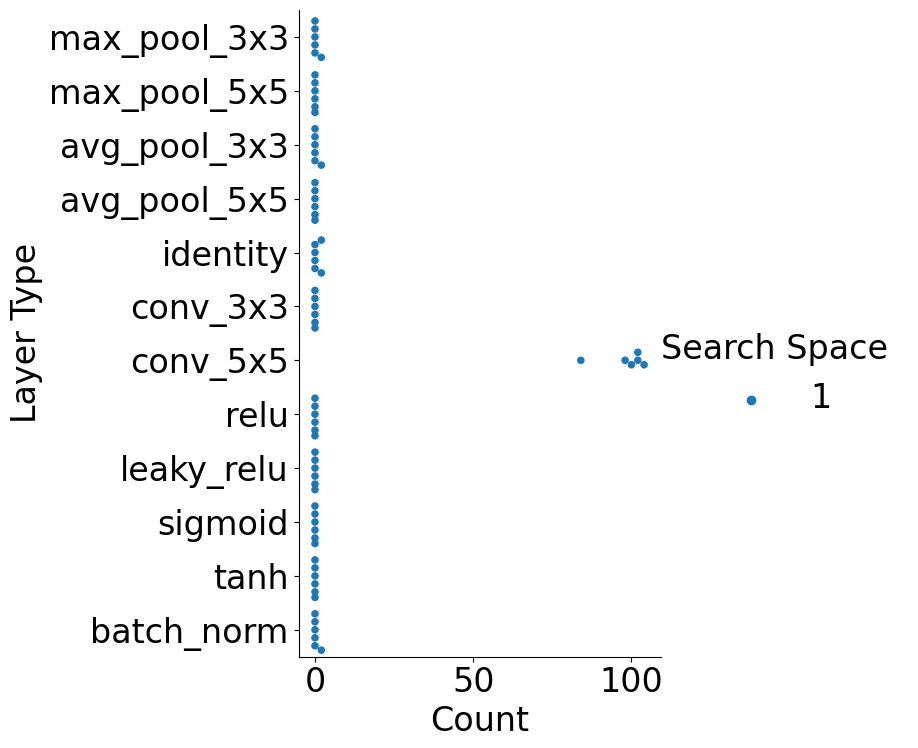}
\includegraphics[width=0.45\columnwidth]{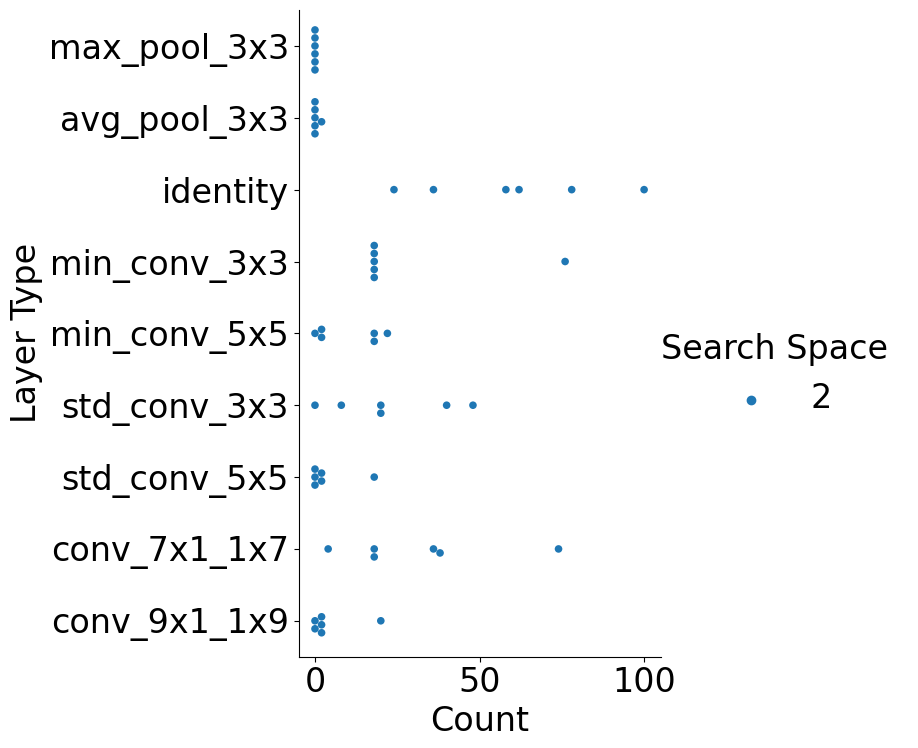}\\
\vspace{0.1cm}
\includegraphics[width=0.45\columnwidth]{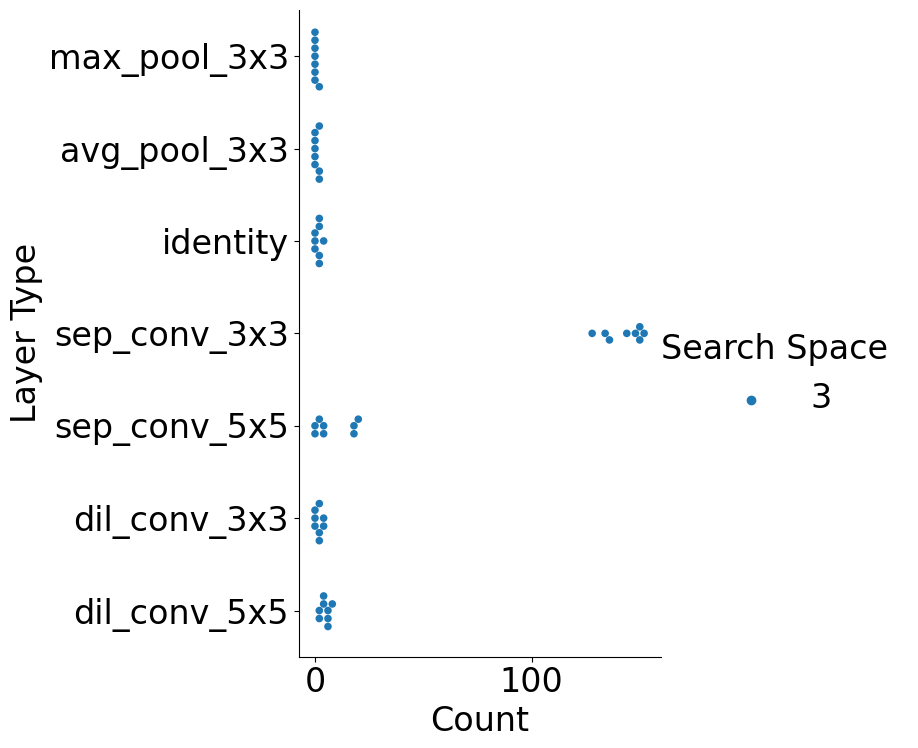}
\includegraphics[width=0.45\columnwidth]{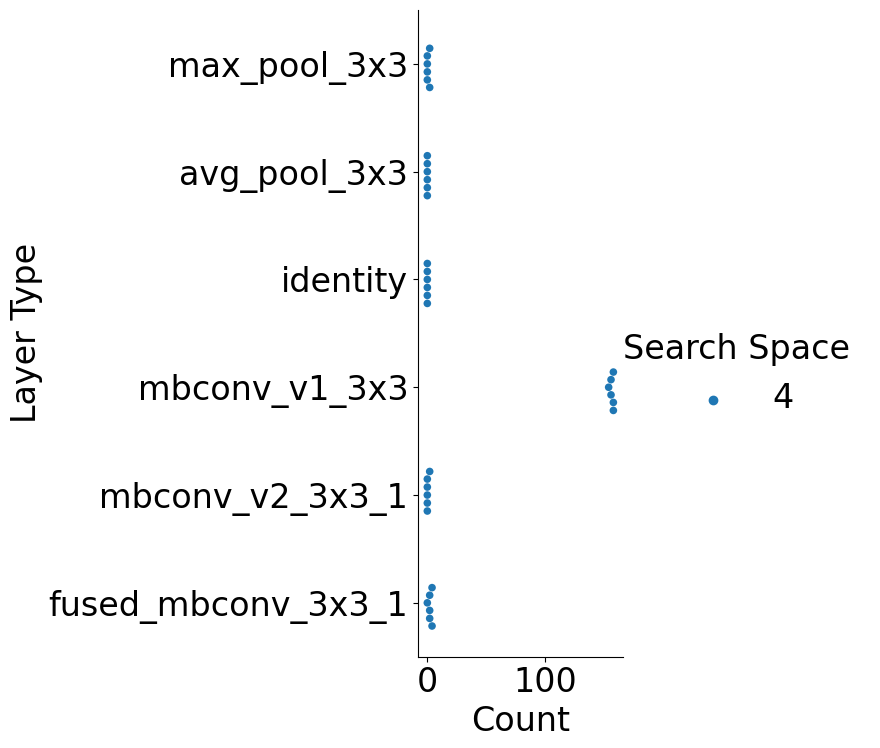}
\caption{ICDARTS Alternative Operation Search Spaces Per Network Layer Type Frequencies. The details of each search space are listed in Table \ref{tab-search-spaces}.}
\label{fig-expanded-search-catplot-layers}
\end{center}
\end{figure*}

\begin{figure*}[t]
  
\begin{center}
\centerline{
\includegraphics[width=0.5\columnwidth]{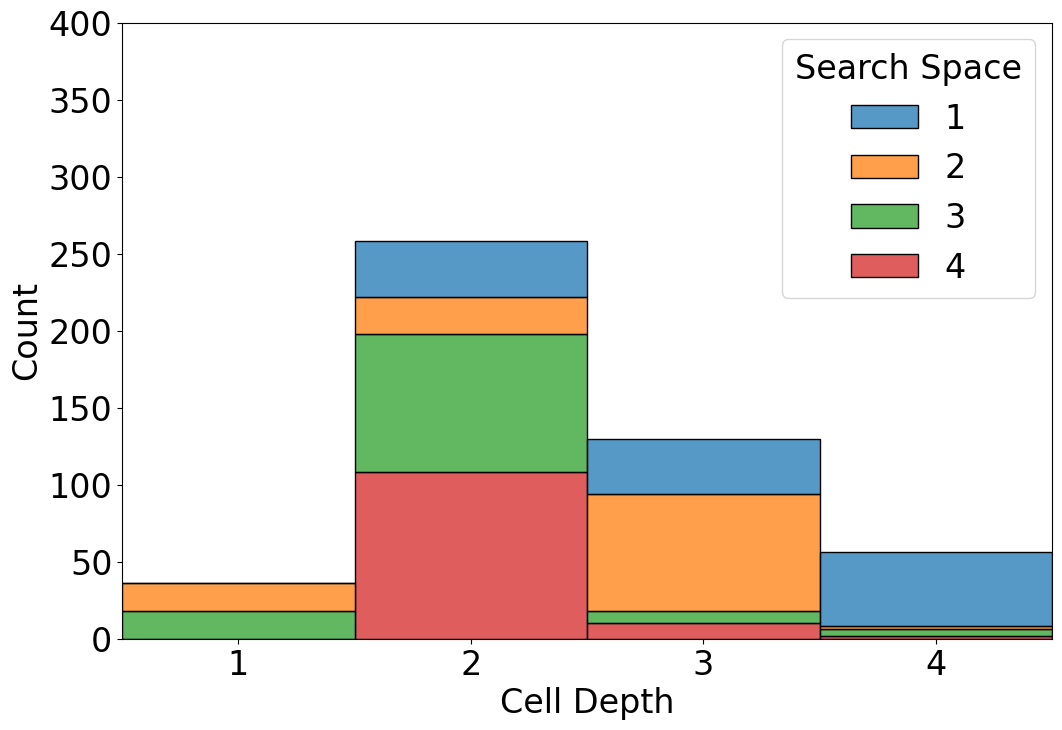}}
\caption{ICDARTS Alternative Search Spaces Per Network Cell Depth Frequencies. The details of each search space are listed in Table \ref{tab-search-spaces}.}
\label{fig-expanded-search-depths}
\end{center}
\end{figure*}

The results from running ICDARTS with different operation search spaces are shown in Table \ref{tab-search-spaces} and Figures \ref{fig-expanded-search-catplot-layers} and \ref{fig-expanded-search-depths}. Search space 2 achieved the highest average retraining accuracy despite using simpler operations than the default search space 3, yet had slightly more variation in accuracy. This outcome might be explained by ICDARTS' tendency to favor slightly deeper cell architectures with this search space than with search space 3 or by this search space's diverse, yet competitive selection of operation choices that appears to balance operation complexity and performance, resulting in the search algorithm preferring cells composed of a diverse selection of simpler operations. By contrast, the cells produced by the other search spaces tended to favor one operation choice above all others. The networks produced with search space 1, favored the convolution with kernel size 5 operation and slightly deeper cell architectures than those produced by the original search space, although these architectures yielded the worst accuracies of any search space. Search space 4, which used the most complex operations and favored the MBConvV1 operation, produced the second lowest accuracies. However, this search space's poor performance and inference latency could be explained by the limitations placed on this search space due to the high computational demands of its operations.

\begin{table}[h]
\caption{ICDARTS Expanded Search CIFAR-10 Retraining Test Set Accuracies and Inference Latencies}
\begin{center}
\begin{tabular}{|c|c|c|}
\hline
\textbf{Search Space}& \textbf{Retraining Accuracy} & \textbf{Inference Latency (s/batch)}\\ 
\hline
1* & 95.08 (0.39) & 0.11 (0.02)\\
\hline
2* & 97.29 (0.29) & 0.10 (0.02)\\
\hline
3 & 97.13 (0.14) & 0.10 (0.01)\\
\hline
4* & 96.55 (0.21) & 0.21 (0.01)\\
\hline
Dynamic* & 97.25 (0.06) & 0.10 (0.01)\\
\hline
\multicolumn{3}{l}{Note: Results from 6 runs are indicated by *.}\\
\end{tabular}
\label{tab-expanded-ss-results}
\end{center}
\end{table}

\begin{figure*}[t]

\begin{center}
\centerline{
\includegraphics[width=0.45\columnwidth]{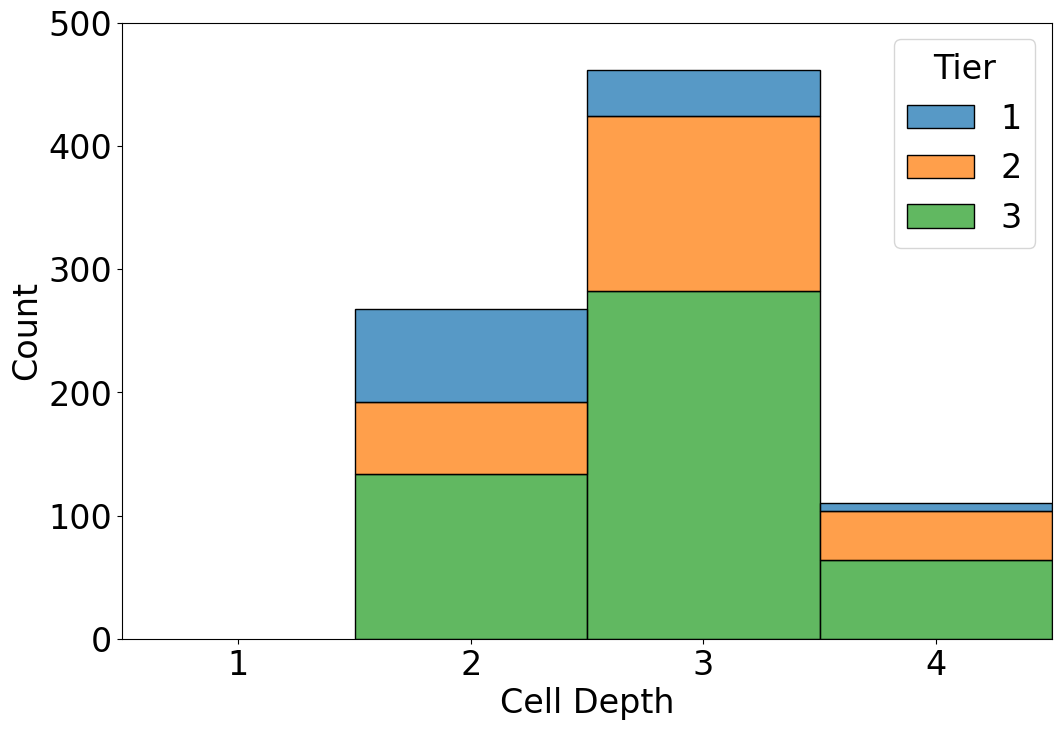}
}
\caption{ICDARTS Dynamic Search Space Algorithm Tiers Per Network Cell Depth Frequencies}
\label{fig-dynamic-cell-depths}
\end{center}
\end{figure*}

\begin{figure*}[t]
  
\begin{center}
\centerline{
\includegraphics[width=0.9\columnwidth]{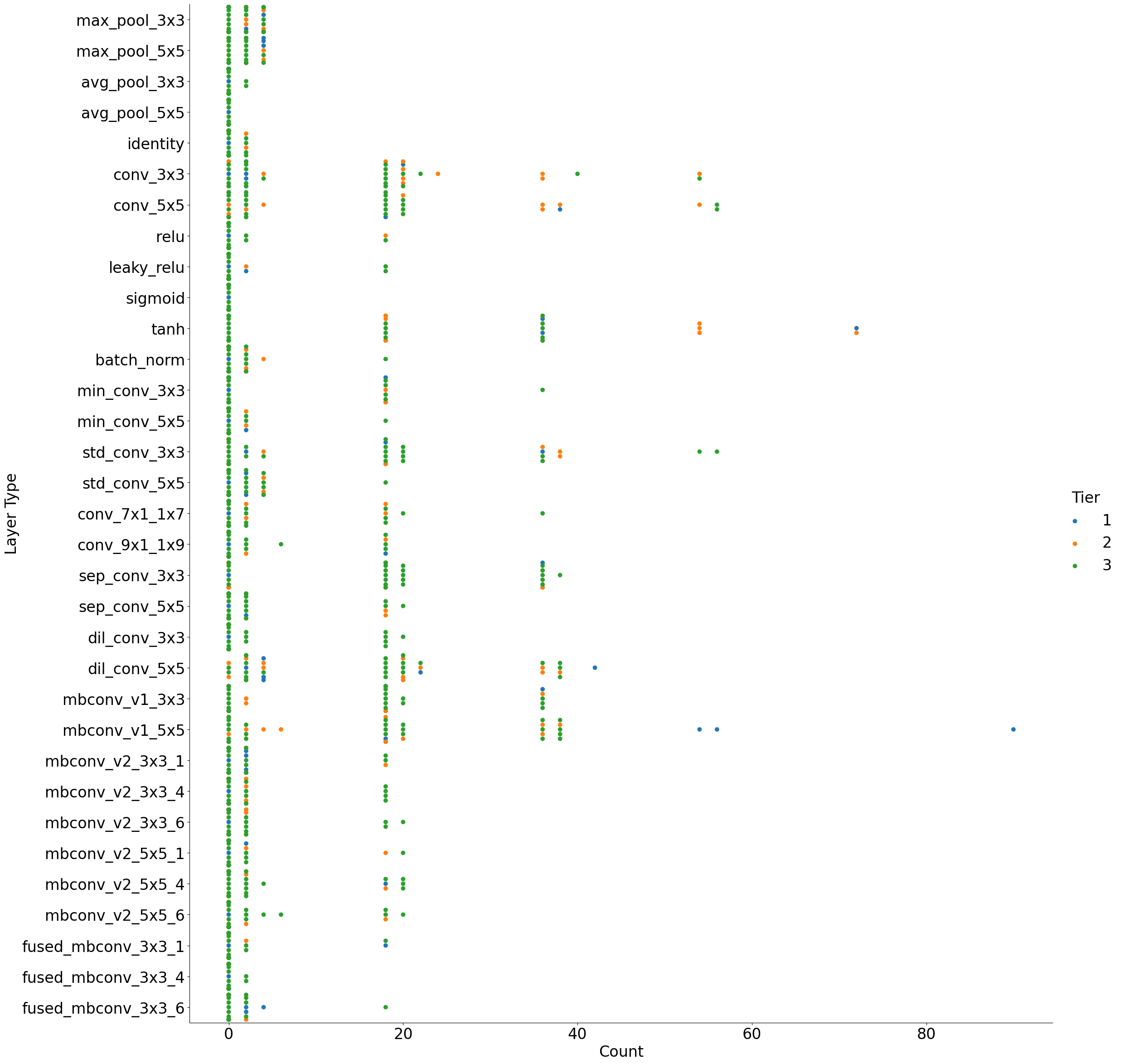}
}
\caption{ICDARTS Dynamic Search Space Algorithm Tiers Per Network Layer Type Frequencies}
\label{fig-dynamic-catplot}
\end{center}
\end{figure*}

The average accuracy and inference latency of ICDARTS using the dynamic search space algorithm are listed alongside those of the other search spaces in Table \ref{tab-expanded-ss-results}. Additional results from each tier of the search algorithm are displayed in Figures  \ref{fig-dynamic-catplot} and \ref{fig-dynamic-cell-depths}. The tournament-style dynamic search space algorithm applied to the master operation space of Table \ref{tab-dynamic-search} achieved an accuracy second only to that of search space 2, but with one of lowest standard deviations of any experiment performed on ICDARTS and an inference latency on par with that of ICDARTS given the 2nd and 3rd search spaces. As shown in Figure \ref{fig-dynamic-catplot}, the cells produced by the final tier of the algorithm most favored the tanh, MBConv, and simple, standard, and separable convolution operations. All of these operations were among the top in their derivative search spaces, with the exception of the tanh activation and the standard convolutions, possibly due to the ability of these operations to pair well with others. Figure \ref{fig-dynamic-cell-depths} reveals that deeper cell architectures were favored in the lower tiers, in which the cell operations varied most due to the operation spaces being randomly spawned for each cell at this level. However, as the architectures converged towards a selection of the most optimal layer operations in the higher tiers, shallower architectures were favored. This outcome suggests that, when given a large and diverse search space of candidate operations, the effectiveness of layer choices may be more important than cell depth for designing an optimal network architecture. It may also explain how this search space produced networks with relatively low inference latencies and suggests that this algorithm may prove helpful for discovering networks with both low inference latencies and high generalization abilities.

\begin{table}[h]
\caption{ICDARTS Alternative Cells CIFAR-10 Retraining Test Set Accuracies and Inference Latencies}
\begin{center}
\begin{tabular}{|c|c|c|}
\hline
\textbf{Ablation}& \textbf{Retraining Accuracy} & \textbf{Inference Latency (s/batch)}\\ 
\hline
DARTS Cells & 97.13 (0.14) & 0.10 (0.01)\\
\hline
I-DARTS Cells & 96.87 (0.27) & 0.10 (0.01)\\
\hline
XDARTS Cells & 97.21 (0.09) & 0.17 (0.01)\\
\hline
\end{tabular}
\label{tab-alt-cells-results}
\end{center}
\end{table}

\begin{figure*}[t]
  
\begin{center}
\centerline{
\includegraphics[width=0.45\columnwidth]{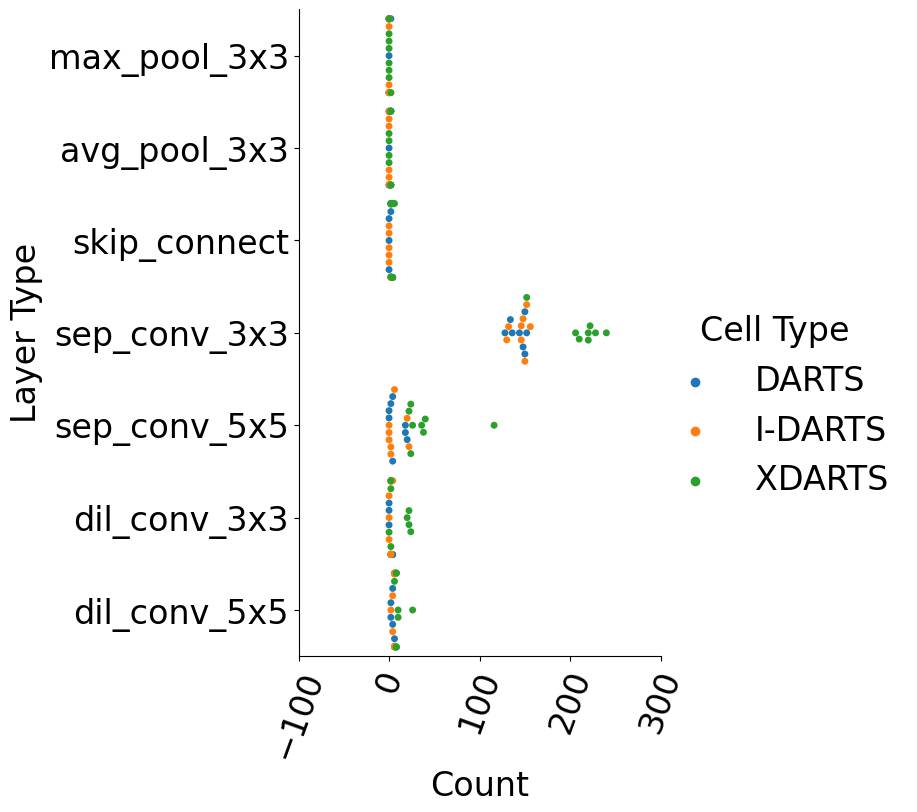}
\includegraphics[width=0.45\columnwidth]{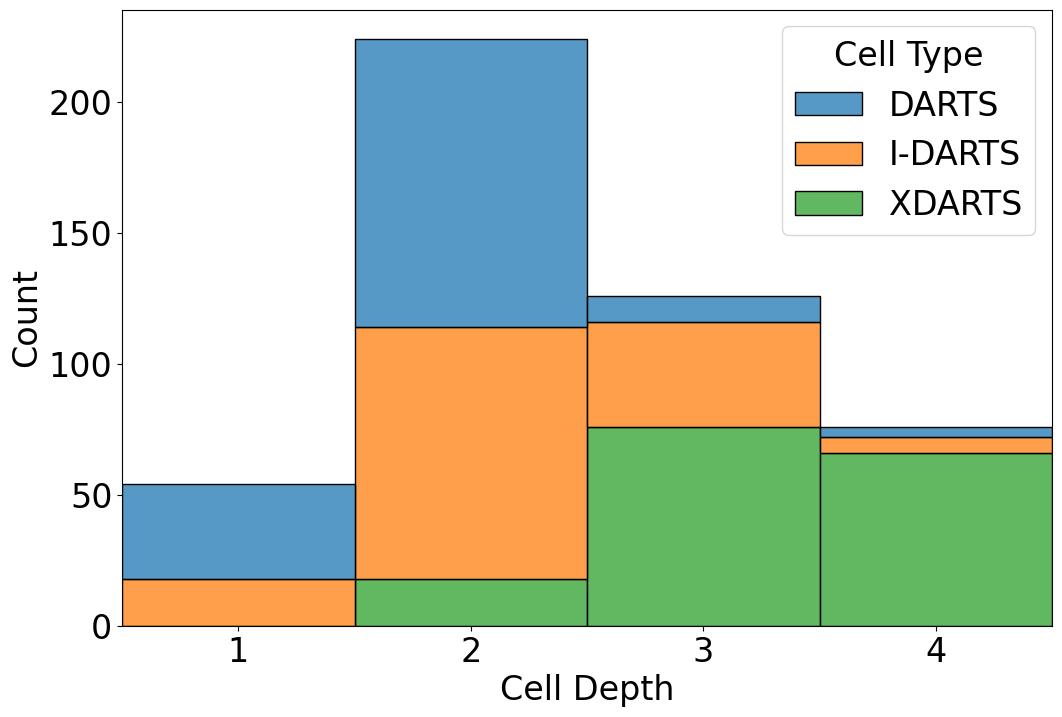}
}
\caption{ICDARTS Alternative Cell Discretization Approaches: Per Network Layer Type Frequencies (left) and Cell Depth Frequencies (right)}
\label{fig-alternative-cells-catplot}
\end{center}
\end{figure*}

Table \ref{tab-alt-cells-results} and Figure \ref{fig-alternative-cells-catplot} show that the I-DARTS cells produced results significantly worse than the IDARTS cells in terms of accuracy and stability, while the opposite was the case with the XDARTS cells. Since the depths and operation choices of the searched DARTS and I-DARTS cells are similar, with the I-DARTS cells even tending to be slightly deeper than the DARTS cells, the reason for the poor performance and instability of I-DARTS cells is not immediately apparent. However, we believe that the poor performance of the I-DARTS cells is due to the softmax being applied across all incoming edges to a node, the number of which increased based on the depth of the node in the cell graph, rather than to a constant number of edges, as was the protocol with the DARTS cells. The results from the XDARTS cells further support this interpretation and show that the stability and performance reductions caused by applying softmax functions over increasing numbers of edges can be rectified by allowing the selection of additional edges based on a node's depth in the cell. Although the XDARTS cells produced significantly more stable and accurate results than the DARTS cells, a severe drawback was the high latencies of these networks. To address this problem, we are working on a multi-objective version of ICDARTS that optimizes networks for latency and performance.

\section{Conclusion}
In this work, we presented improvements to the original CDARTS algorithm and search space, yielding results with superior stability and generalization performance. The optimized networks produced by our initial set of improvements obtained similar accuracies to those produced by CDARTS while maintaining a minor variance (typically at least 2$\times$ smaller). Additionally, we evaluated alternative \textit{zero} operation schemes and demonstrated that the best performance and stability results could be achieved by completely excluding the \textit{zero} operation from the search phase. Alternatively, if it is desired to include \textit{zero} operations in the discretized networks, we found that this could be accomplished by replacing the \textit{zero} operation with a \textit{random} operation during the search phase.
We then conducted ablation studies on the improved algorithm and its network template. The results showed that improved accuracies could be obtained by removing pooling operations from its search space. The search, inference, and retraining latencies were minimized most by eliminating the separable convolution operations and the reduce cells, although both changes negatively impacted the retraining accuracies. The algorithmic ablation study generally showed that each change offered stability improvements over the original algorithm.
Finally, we experimented with expanding the search space of our algorithm by introducing new operation search spaces and new search cell discretization methods that increased the diversity of cells that could be discovered. These experiments spawned the creation of a novel algorithm for efficiently traversing large search spaces using the improved search method.

\acks{This research used resources of the Compute and Data Environment for Science (CADES) at the Oak Ridge National Laboratory, which is supported by the Office of Science of the U.S. Department of Energy under Contract No. DE-AC05-00OR22725

This research used resources of the Oak Ridge Leadership Computing Facility at the Oak Ridge National Laboratory, which is supported by the Office of Science of the U.S. Department of Energy under Contract No. DE-AC05-00OR22725.
}


\newpage

\vskip 0.2in
\bibliography{main}

\end{document}